%% file: main.tex
\documentclass[letterpaper, 10 pt, conference]{ieeeconf}  
\IEEEoverridecommandlockouts                              
\usepackage[utf8]{inputenc}
\usepackage[T1]{fontenc}
\usepackage{paralist}
\usepackage{pifont}

\usepackage{graphics} 
\usepackage{epsfig} 
\usepackage{mathptmx} 
\usepackage{mathptmx} 
\usepackage{amsmath} 
\usepackage{amssymb}  

\usepackage{multirow}
\usepackage{tabulary}
\usepackage[table]{xcolor} 
\usepackage{cite}

\definecolor{blue}{rgb}{0,0,1}

\definecolor{green}{rgb}{0,0.5,0}

\definecolor{orange}{rgb}{1,0.2,0}

\usepackage{caption}
\let\oldtwocolumn\twocolumn
\renewcommand\twocolumn[1][]{%
    \oldtwocolumn[{#1}{
    \begin{center}
    \vskip-5.5ex
        \centering
        \includegraphics[width=0.75\textwidth]{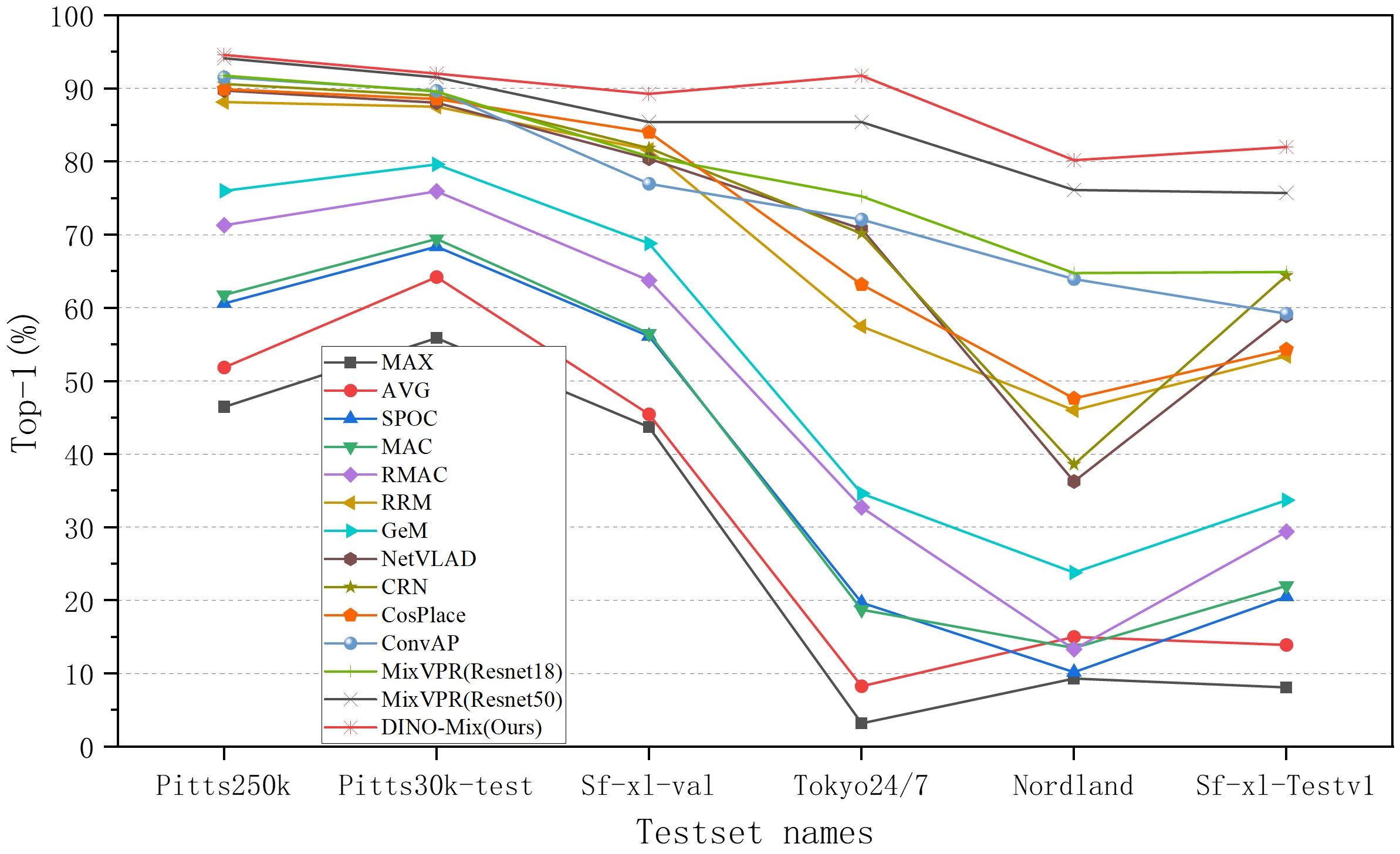}
        \end{center}
        \vskip-1em
        \captionof{figure} {
\textbf{\emph{Comparison of DINO-Mix and other VPR methods in terms of Top-1 accuracy for different test sets}} }
        \label{fig:teaser}
    }]
}

\makeatletter
\let\NAT@parse\undefined
\makeatother
\usepackage[colorlinks]{hyperref}
\hypersetup{colorlinks=true,linkcolor=red,citecolor=blue,urlcolor=magenta}

\definecolor{revised_color_SH}{HTML}{007FFF}

\usepackage[capitalize]{cleveref}
\crefname{section}{Sec.}{Secs.}
\Crefname{section}{Section}{Sections}
\Crefname{table}{Table}{Tables}
\crefname{table}{Tab.}{Tabs.}

\usepackage{setspace}

\title{\LARGE \bf DINO-Mix: Enhancing Visual Place Recognition with Foundational Vision Model and Feature Mixing}

\author{Gaoshuang Huang, Yang Zhou$^{*}$, Xiaofei Hu$^{}$, Chenglong Zhang$^{}$, Luying Zhao$^{}$, Wenjian Gan, and Mingbo Hou  
\thanks{This paper is supported in part by the National Natural Science Foundation of China (NSFC) under Grant No.42001338.}
\thanks{* Corresponding author}
\thanks{All the authors are with the Institute of Geospatial Information, PLA Strategic Support Force Information Engineering University, Zheng Zhou, China.E-mail: huanggaoshuang123@163.com, zhouyang3d@163.com, huxiaofeicn@163.com, 
z1204149693@163.com, 2014202130068@whu.edu.cn, 14737117985@163.com, hobefrank@163.com.}}

\begin{document}

\maketitle
\thispagestyle{plain} 

\begin{abstract}
\input{Tex_content/abstract}
\end{abstract}

\section{INTRODUCTION}
\label{sec:introduction}
\input{Tex_content/intro}

\section{RELATED WORK}
\label{sec:related work}
\input{Tex_content/related_work}

\section{METHODOLOGY}
\label{sec:methodology}
\input{Tex_content/methodology}

\section{EXPERIMENTS}
\label{sec:experiments}
\input{Tex_content/experiments}

\section{QUALITATIVE RESULTS}
\label{sec:Qualitative results}
\input{Tex_content/qualitative_results}

\section{CONCLUSIONS}
\label{sec:conclusions}
\input{Tex_content/conclusion}


\input{main.bbl}

\end{document}

%% file: Tex_content/abstract.tex
In the vast expanse of cyberspace, a plethora of publicly available images exist. Utilizing visual place recognition (VPR) technology to ascertain the geographical location of publicly available images is a pressing issue for real-world VPR applications. Although most current VPR methods achieve favorable results under ideal conditions, their performance in complex environments, characterized by lighting variations, seasonal changes, and occlusions caused by moving objects, is generally unsatisfactory. Therefore, obtaining efficient and robust image feature descriptors even in complex environments is a pressing issue in VPR applications. In this study, we utilize the DINOv2 model as the backbone network for trimming and fine-tuning to extract robust image features. We propose a novel VPR architecture called DINO-Mix, which combines a foundational vision model with feature aggregation. This architecture relies on the powerful image feature extraction capabilities of foundational vision models. We employ an MLP-Mixer-based mix module to aggregate image features, resulting in globally robust and generalizable descriptors that enable high-precision VPR. We experimentally demonstrate that the proposed DINO-Mix architecture significantly outperforms current state-of-the-art (SOTA) methods. In test sets having lighting variations, seasonal changes, and occlusions (Tokyo24/7, Nordland, SF-XL-Testv1), our proposed DINO-Mix architecture achieved Top-1 accuracy rates of $91.75\%$, $80.18\%$, and $82\%$, respectively. Compared with SOTA methods, our architecture exhibited an average accuracy improvement of $5.14\%$. To further evaluate the performance of DINO-Mix, we compared it with other SOTA methods using representative image retrieval case studies. Our analysis revealed that DINO-Mix outperforms its competitors in terms of VPR performance. Furthermore, we visualized the attention maps of DINO-Mix and other methods to provide a more intuitive understanding of their respective strengths. These visualizations serve as compelling evidence of the superiority of the DINO-Mix framework in this domain. Code is available at \url{https://github.com/GaoShuang98/DINO-Mix}.

%% file: Tex_content/intro.tex
Visual place recognition (VPR), also known as image geo-localization (IG) or visual geo-localization (VG), has been extensively applied in various fields, such as cyberspace mapping, intelligence gathering, image target localization, autonomous driving, and outdoor user localization. Currently, most VPR studies focus on image retrieval in urban scenarios. However, images captured in urban environments may have varying shooting angles, temporal lighting changes, seasonal variations, occlusions, and similar repetitive textures. These conditions can pose significant difficulty for achieving high-precision image retrieval. Therefore, extracting robust and generalizable image feature descriptors for accurate image retrieval is a critical issue.

In previous approaches to VPR, handcrafted SIFT~\cite{lowe_distinctive_2004}, HOG~\cite{dalal_histograms_2005}, SURF~\cite{leonardis_surf_2006}, ORB~\cite{rublee_orb_2011}, and other techniques were used to extract features from images. These features were then aggregated using methods such as bag-of-words (BoW)~\cite{tang_learning_2012}, Fisher Vector (FV)~\cite{jegou_aggregating_2010}, and Vector of Locally Aggregated Descriptors (VLAD)~\cite{jegou_aggregating_2012} to obtain image descriptors for image retrieval, enabling image geo-localization. More recently, deep-learning techniques have emerged as mainstream approaches for extracting image features. Compared with handcrafted features, these methods significantly improve VPR accuracy. Examples include NetVLAD~\cite{relja_netvlad_2018}, which combines convolutional neural networks (CNNs) with VLAD, and other variants that incorporate attention, semantics, context, and multiscale features. Other methods based on the generalized mean (GeM)~\cite{radenovic_fine-tuning_2019}, such as CosPlace~\cite{berton_rethinking_2022}, and those utilizing fully-connected multilayer perceptron (MLP)-based feature aggregation, such as MixVPR~\cite{ali-bey_mixvpr_2023}, have been proposed. However, in practical testing, the accuracy of these methods for image geo-localization has been suboptimal under challenging conditions, such as varying shooting angles, temporal lighting changes, seasonal variations, and occlusions. 

 The rapid development of foundational visual models has enabled the generation of universal visual features from images~\cite{oquab_dinov2_2023}. By training on billions of data points, foundational visual models can extract image features that are more generalizable and robust than those extracted by conventional models. They can effectively handle the challenging conditions encountered in practice. Therefore, incorporating foundational visual models into VPR is a promising approach.

Considering the aforementioned challenges, this study proposes a method based on the DINOv2~\cite{oquab_dinov2_2023} model, called DINO-Mix, which combines foundational visual models with feature aggregation. This architecture possesses exceptional discriminative power. Efficient and robust image features suitable for image geo-localization are extracted by fine-tuning and trimming. Furthermore, it utilizes a feature mixer~\cite{tolstikhin_mlp-mixer_2021} module to aggregate image features, resulting in a global feature descriptor vector. DINO-Mix is experimentally demonstrated to achieve superior test accuracy on multiple benchmarks, surpassing state-of-the-art (SOTA) methods.

The remainder of this paper is organized as follows. In  \ref{sec:related work}, we summarize previous relevant research on image geo-localization. In \ref{sec:methodology}, the DINO-Mix method is introduced. In \ref{sec:experiments}, we provide details on the training set, testing set, training, and evaluation parameters used in our experiments. The proposed method is compared with existing methods in terms of accuracy, and ablation experiments are conducted. In \ref{sec:Qualitative results}, We qualitatively demonstrate the state-of-the-art of DINO-Mix architecture by comparing DINO-Mix with other VPR methods through a typical VPR example and visualizing the corresponding attention map. Finally, our conclusions are presented in \ref{sec:conclusions}.

%% file: Tex_content/related_work.tex
By retrieving the most similar image from the image database, the geographical location of the retrieved image can be used as the location of the target image~\cite{masone_survey_2021}. In recent years, numerous researchers have made significant contributions to the field of image retrieval for VPR. The features used in image retrieval can be broadly categorized into handcrafted and deep features. Zhang and Kosecka~\cite{zhang_image_2007} first extracted scale-invariant feature transform (SIFT) features from images to establish an image feature database. They performed a brute-force global search of the database and validated and ranked the top five candidate images using the random sample consensus (RANSAC)~\cite{martin_a_fischler_random_1981} algorithm. Finally, the geographical location of the target image was obtained by triangulating the top three images. Zamir and Shah~\cite{zamir_accurate_2010} extracted SIFT feature vectors from images to build a database and employed a nearest-neighbor tree search to improve the retrieval efficiency. Zamir and Shah~\cite{zamir_gps-tag_2014} further improved the nearest-neighbor matching technique by pruning outliers, and applied the generalized minimum clique problem (GMCP) in conjunction with approximate feature matching. This resulted in a $5\%$ improvement in the localization accuracy compared to their previous work~\cite{zamir_accurate_2010}. The advantages of using handcrafted features for VPR are their simplicity and strong interpretability. However, these methods tend to have high redundancy, require dimensionality reduction, are susceptible to environmental changes, and generally have low accuracy.

Deep features are extracted by neural networks with modules such as convolutional layers and attention mechanisms. These features often outperform handcrafted features owing to their strong expressive power, ability to freely define feature dimensions, and flexibility in designing neural network frameworks. Noh et al.~\cite{noh_large-scale_2017} proposed a deep local feature (DELF) descriptor and an attention mechanism for keypoint selection to identify semantic local features. Ng et al.~\cite{ng_solar_2020} introduced a global descriptor called Second-Order Loss and Attention for image Retrieval (SOLAR) that utilizes spatial attention and descriptor similarity to perform large-scale image retrieval using second-order information. Chu et al.~\cite{chu_grid_2020} constructed a CNN to extract dense features, embedded an attention module within the network to score features, and proposed a grid feature point selection (GFS) method to reduce the number of image features. Chu et al.~\cite{chu_street_2020} combined deep features with handcrafted features, extracted average pooling features from the intermediate layers of a CNN for retrieval on street-view datasets, and used SIFT to re-rank them. Yan~\cite{yan_hierarchical_2021} extracted hierarchical feature maps from CNNs and organically fused them for image feature representation. Chu et al.~\cite{chu_news_2022} employed a CNN with a HOW module~\cite{tolias_learning_2020} to extract local image features, aggregated them into a feature vector using VLAD, used the aggregated selective match kernel (ASMK), and estimated the geographical location of the query image using kernel density prediction (KDP).

To address environmental factors, Mishkin et al.~\cite{mishkin_place_2015} employed a BoW method with multiple detectors, descriptors, and adaptive thresholds. Relja et al.~\cite{relja_netvlad_2018} designed a trainable NetVLAD layer inspired by VLAD, which provides a pooling mechanism that can be integrated into other CNN structures. In addition, variants of NetVLAD have been proposed, such as CRN~\cite{kim_learned_2017}, SPE-VLAD~\cite{yu_spatial_2020}, MultiRes-NetVLAD~\cite{khaliq_multires-netvlad_2022}, SARE~\cite{liu_stochastic_2019}, and SFRS~\cite{ge_self-supervising_2020}. Ali-bey et al. proposed ConvAP~\cite{ali-bey_gsv-cities_2022}, which combines 1×1 convolutions with adaptive mean pooling to encode local features.

%% file: Tex_content/methodology.tex
\subsection{Proposed architecture}
\label{proposed architecture}

\begin{figure*}[!t]
\renewcommand{\thefigure}{2} 
    \centering
    \includegraphics[width=0.6\linewidth]{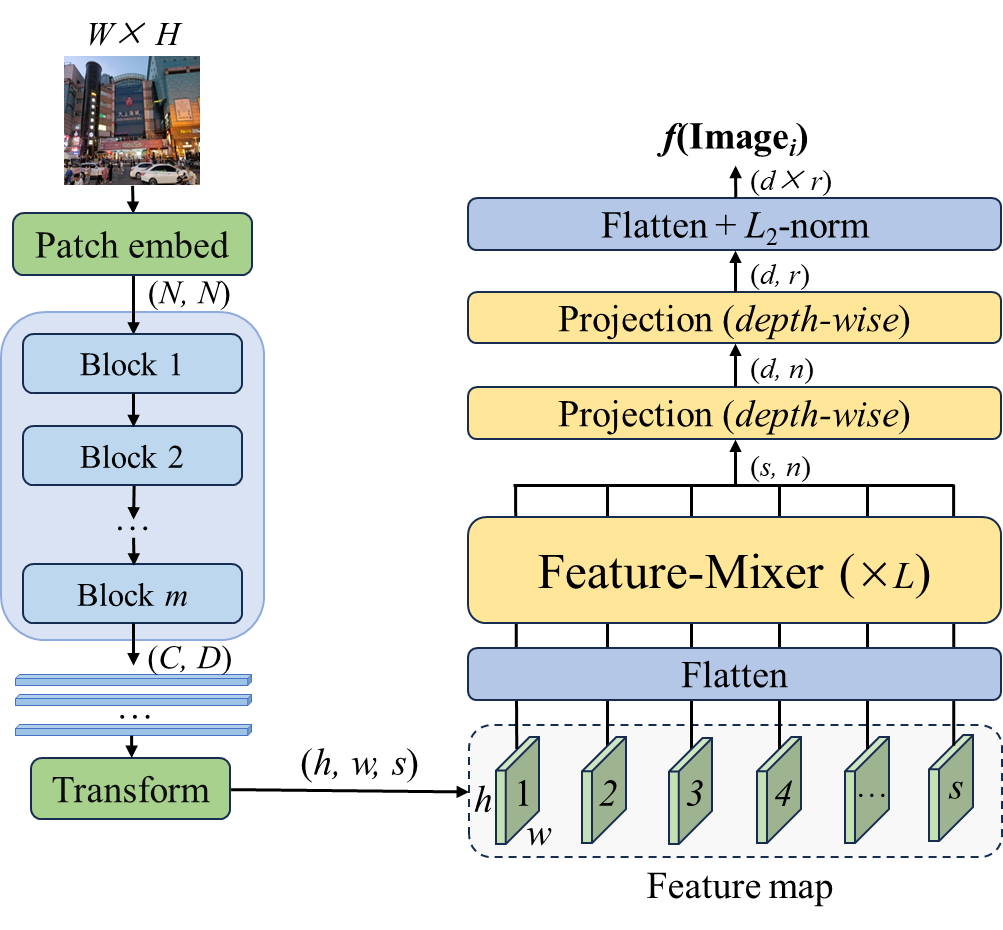}
    \vspace{0em}
    \caption{\emph{\textbf{The visual place recognition structure of DINO-Mix.}}}
    \label{fig:DINO-Mix_architecture}
\end{figure*}

We propose an image-geolocation framework that integrates the foundational Vision model with feature aggregation. The proposed framework utilizes the truncated DINOv2~\cite{oquab_dinov2_2023} model as the backbone network. Owing to its exceptional image understanding capability, DINOv2 is well-suited for various downstream tasks. Therefore, we pre-trained the DINOv2 model as the primary network for image feature extraction and employed an efficient and lightweight Mixer module to aggregate the obtained image features. The DINO-Mix visual geolocation architecture is illustrated in Fig.~\ref{fig:DINO-Mix_architecture}.

We modified the DINOv2 model by removing its layer norms and head modules, which were subsequently used as the backbone network. Furthermore, to maximize the pre-trained parameter benefits of the DINOv2 model for image understanding, we used the output from the last layer of the Vision Transformer (ViT) blocks~\cite{dosovitskiy_image_2021} as the input to the Mixer module. Given that the output of the modified ViT block module is a feature matrix of size C × D (channels × feature vector length), we transform it into s feature maps of size h × w, as expressed by Equation \ref{equation6}. These transformed feature maps serve as inputs for the mix module.

\begin{equation}
\label{equation6}
    \begin{cases}
        D = s \\
        C = hw
    \end{cases}
\end{equation}

where $D$ represents the length of the feature vector output by the backbone network, $C$ denotes the number of channels in the output of the backbone network, $h$ and $w$ are the height and width of the feature map, respectively, and $s$ is the number of feature maps.

\subsection{Foundational vision model: DINOv2}
\label{Large Vision Model:DINOv2}

Foundational vision models (FVMs) are typically constructed using structures such as CNNs or Transformers. These models often have parameters on the order of tens to hundreds of millions, giving them a greater representational capacity than smaller models. In addition, because of the use of larger and more diverse datasets during training, FVMs can learn more features and have better generalization capabilities.

\begin{figure}[!t]
\renewcommand{\thefigure}{3}    %
   \centering
   \includegraphics[width=0.7\linewidth]{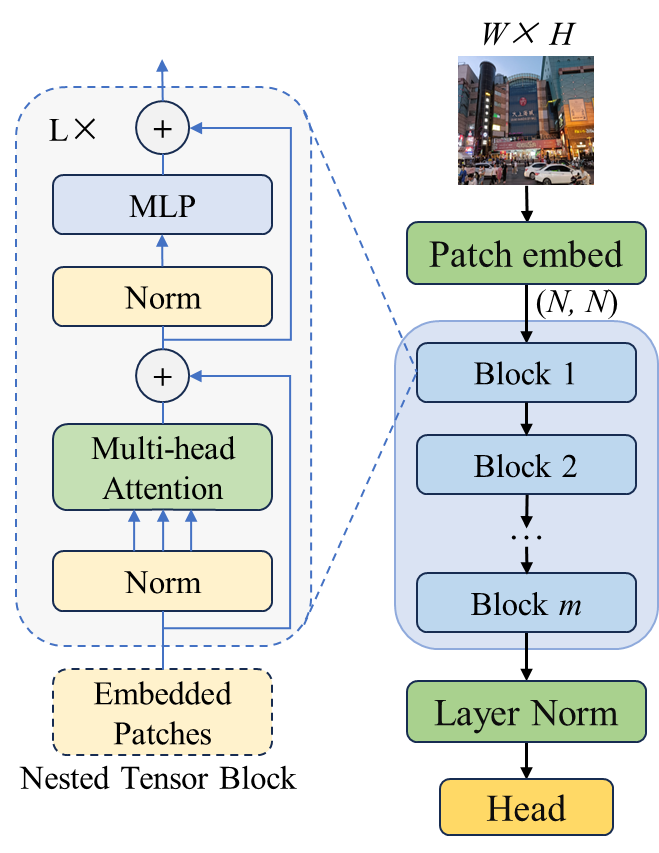}
   \vspace{0em}
   \caption{\textbf{\emph{The structural diagram of the DINOv2 model}}. }
   \label{fig:dinov2_structure}
  \vspace{-1.5em}
\end{figure}

DINOv2~\cite{oquab_dinov2_2023} is capable of extracting powerful image features and performs well across various tasks. Compared with Segment Anything~\cite{kirillov_segment_2023}, DINOv2 has a broader scope of application and areas of use. The architecture of the DINOv2 model is illustrated in Fig.~\ref{fig:dinov2_structure}. First, an input image is passed through a patch-embedded module consisting of a two-dimensional (2D) convolutional layer with a kernel size of 14 × 14 and a stride of 14, followed by a normalization layer. This process uniformly outputs patches of size 14 × 14. These patches are then fed into ViT blocks, which vary in number according to the size of the model. The ViT blocks output a feature matrix of size $C$ (number of channels) × $D$ (dimension of the feature vector), which is then normalized by a layer-norm module before being transformed into a feature vector of size 1 × $n$. Finally, the head module can be flexibly selected based on specific image task requirements.

DINOv2 is characterized by several key features: a) it presents a novel approach for training high-performance computer vision models; b) it offers superior performance without the need for fine-tuning; c) it can learn from any image dataset and capture certain features that existing methods struggle with; and d) it leverages knowledge distillation to transfer knowledge from more complex teacher models to smaller student models. Through knowledge distillation, three smaller models were obtained from the ViTg14 model: ViTl14 (large), ViTb14 (base), and ViTs14 (small) (see Tab.~\ref{tab:Four ViT model param}).

\begin{table}[!t]
\renewcommand{\thetable}{1}    %
    \caption{\emph{\textbf{Four ViT model parameters for DINOv2}}}
    \centering
    \begin{tabular}{c p{1.2cm} c p{1cm} c p{1cm} c p{1.1cm} c p{1cm}}
    \hline
    Name & Patch embed & Blocks & Feature dim & Size(MB) \\
    \hline
    ViTs14 & 14 × 14 & 12 & 384 & 86.2 \\
    ViTb14 & 14 × 14 & 24 & 768 & 338.2 \\
    ViTl14 & 14 × 14 & 24 & 1024 & 1189.0 \\
    ViTg14 & 14 × 14 & 40 & 1536 & 4439.5 \\
    \hline
    \end{tabular}
    \label{tab:Four ViT model param}
\end{table}

The primary advantage of DINOv2 is the ability to create a large dataset for model training. This dataset, called LVD-142M, comprises 142 million images and includes ImageNet-22k, ImageNet-1k, Google Landmarks, various fine-grained datasets, and image datasets crawled from the Internet. For model training, an Nvidia A100 40-GB GPU was utilized, with a total of 22k GPU hours dedicated to training the DINOv2-g model.

\subsection{Feature Mixer}
\label{Feature Mixer}

\begin{figure}[!t]
\renewcommand{\thefigure}{4}
    \centering
    \includegraphics[width=0.9\linewidth]{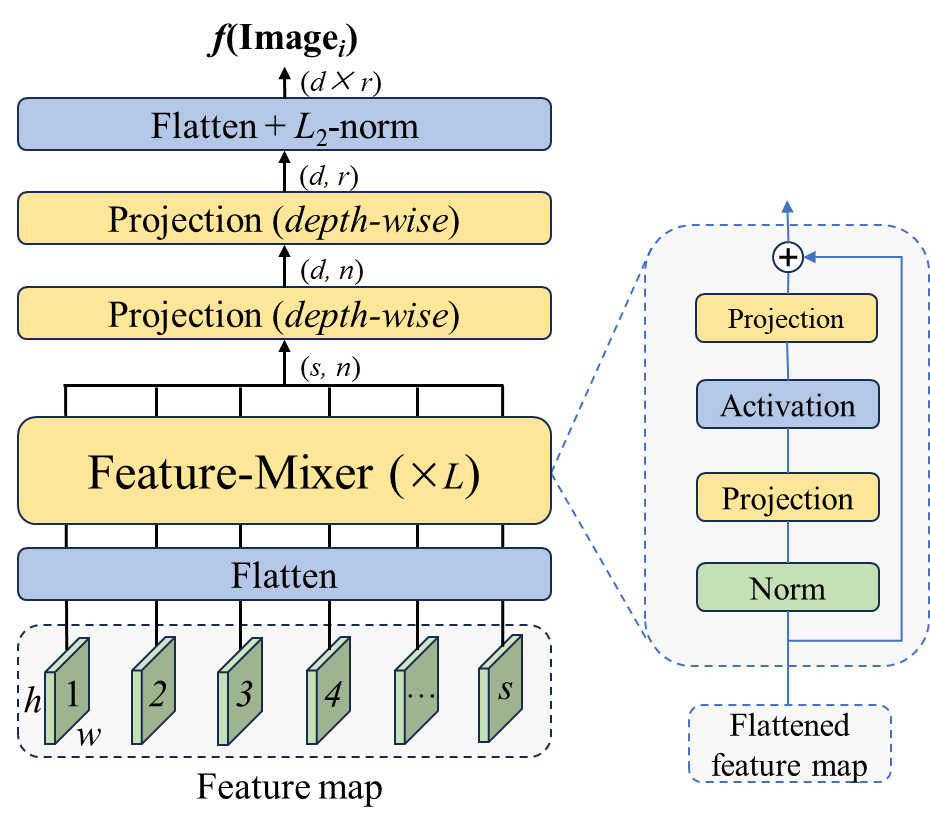}
    \vspace{-1em}
    \caption{\emph{\textbf{The architecture of Mix}}}
    \label{fig:mix_architecture}
\end{figure}

Currently, the most advanced techniques propose shallow aggregation layers that are inserted into very deep pre-trained backbones cropped to the last feature-rich layer. By contrast, Wang et al. proposed TransVPR~\cite{wang_transvpr_2022}, which achieved good results in local feature matching. However, its global representation performance did not surpass that of NetVLAD~\cite{relja_netvlad_2018} or CosPlace~\cite{berton_rethinking_2022}. Recent advancements in isotropic architectures have demonstrated that self-attention is not crucial for ViT. However, Mixer utilizes feature maps extracted from a pre-trained backbone and iteratively merges global relationships into each feature map. This is achieved through an isotropic block stack composed of MLPs, referred to as a feature mixer~\cite{tolstikhin_mlp-mixer_2021}. The effectiveness of Mixer has been demonstrated through several qualitative and quantitative results, demonstrating its high performance and lightweight nature~\cite{ali-bey_mixvpr_2023}; the architecture is illustrated in Fig.~\ref{fig:mix_architecture}.

Mixer treats the input feature map $F\in R^{(s\times h\times w)}$ as a set of s 2D features, each of size $h\times w$, as expressed by Equation \ref{equation1}:

\begin{equation}
\label{equation1}
    F=\{X^{i} \}, i=\{1,…,s\}
\end{equation}

where $X^i$ corresponds to the $i$th activation map in the feature map F. Secondly, each 2D feature map $X^i$ is expanded into a 1D vector representation, resulting in a flattened feature map $F\in R^{(s\times n)}$, where $n=h\times w$.

The flattened feature maps are then fed into the feature mixer, which is composed of $L$ MLPs with the same structure, as shown in Fig.\ref{fig:mix_architecture}. The feature mixer takes the flattened feature map ensemble as input and successively incorporates spatial global relationships into each ${X^i}\in F$ as per Equation \ref{equation2}:

\begin{equation}
\label{equation2}
    {X^i}\gets{W_2}{\big(}\sigma({W_1}{X^i} ){\big)}+{X^i},i=\{1,…,s\}
\end{equation}

where $W_1$ and $W_2$ are the weights of the two fully-connected layers that make up the MLP, and $\sigma$ is the ReLU nonlinear activation function.

For $F\in R^{(s\times n)}$, the feature mixer, owing to its isotropy architecture, produces an output $"Z"\in R^{(s\times n)}$ with the same shape and feeds it into the second feature mixer block, and so on, until $L$ consecutive blocks have been traversed, as per Equation \ref{equation3}:

\begin{equation}
\label{equation3}
   Z={FM_L}{\Big(}FM_{(L-1)}{\big(}…{FM_1} (F){\big)}{\Big)}
\end{equation}

where $Z$ and the feature map F have the same dimensions. To control the dimensions of the final global descriptor, two fully-connected layers are used to successively transform the channel and row dimensions. First, a depth projection is used to map $Z$ from $R^{(s\times n)}$ to $R^{(d\times n)}$, as given by Equation \ref{equation4}:

\begin{equation}
\label{equation4}
    {Z^{'}}={W_d}{\big(}Transpose(Z){\big)}
\end{equation}

where $W_{d}$ denotes the weight of the fully-connected layer. Subsequently, a row-wise projection is used to map the output $Z^{'}$ from $R^{(d\times n)}$ to $R^{(d\times r)}$, as given by Equation \ref{equation5}:

\begin{equation}
\label{equation5}
   O={W_{r}}{\big(}Transpose({Z^{'}}){\big)}
\end{equation}

where $W_r$ denotes the weight of the fully-connected layer. The final output $O$ has dimensions of $d\times r$, which are flattened, and $L2$ is normalized to form a global feature vector.

%% file: Tex_content/experiments.tex
\subsection{Implementation details}
\label{Implementation details}

\textbf{Datasets:} Our model was trained using the GSV-Cities dataset~\cite{ali-bey_gsv-cities_2022}. The following six datasets were employed for evaluation purposes: Pittsburgh250k~\cite{torii_visual_2015} (contains 8k queries and 83k reference images collected from Google Street View and Pittsburgh30k-test), Pittsburgh30k-test~\cite{torii_visual_2015} (a subset of Pittsburgh250k, with 8k queries and 8k reference images), SF-XL-Val dataset~\cite{berton_rethinking_2022}, Tokyo 24/7~\cite{torii_247_2018}, Nordland~\cite{sunderhauf_are_2013}, and SF-XL-Testv1~\cite{berton_rethinking_2022}. The datasets contain extreme variations in lighting, weather, and seasons. Specific information regarding these datasets is presented in Tab.\ref{tab:dataset parameters}.

\begin{table*}[!t]
\renewcommand{\thetable}{2}    %
    \caption{\emph{\textbf{The parameter table of the training dataset and test dataset}}}
    \centering
    \begin{tabular}{c p{1cm} c p{1cm} c p{1cm} c p{0.7cm} c p{1cm}c p{1cm} c p{1cm} c p{1cm} cp{2cm}}
    \hline
    Dataset & Train/val database queries & Test database queries
 & Dataset size (GB) & Database type & Database image size & Urban & Appearance changes\\
    &&&&&&&Season & Day/Night \\
    \hline
    GSV-cities~\cite{ali-bey_gsv-cities_2022}&524701/0&0/0& 21.7& panorama &480×640&\ding{51}&\ding{51}&\ding{51} \\
    Pittsburgh250k~\cite{torii_visual_2015}& 170112/15432&83952/8280&9.5& panorama &300×400&\ding{51}&\ding{55}&\ding{55} \\
    Pittsburgh30k~\cite{torii_visual_2015}&20000/15024&10000/6816&2.0&panorama &480×640&\ding{51}&\ding{55}&\ding{55} \\
    Tokyo 24/7~\cite{torii_247_2018}&0/0&75984/315&4.2&panorama&480×640&\ding{51}&\ding{55}&\ding{51} \\
    Nordland~\cite{sunderhauf_are_2013}& 0/0&27592/27592&1.3&font-view&360×640&\ding{55}&\ding{51}&\ding{55} \\
    SF-XL-Val~\cite{berton_rethinking_2022}& 0/0&8015/7993&0.7&panorama &512×512&\ding{51}&\ding{55}&\ding{55} \\
    SF-XL-Testv1~\cite{berton_rethinking_2022}& 0/0& 27191/1000&1.3&panorama&512×512&\ding{51}&\ding{55}&\ding{51} \\
    \hline
    \end{tabular}
    \label{tab:dataset parameters}
\end{table*}

\textbf{Architecture:} We employed the PyTorch deep-learning framework to implement DINO-Mix. To enable a fair comparison with other methods in terms of accuracy, we conducted precision tests on various VPR frameworks such as NetVLAD, GeM, ConvAP, CosPlace, and MixVPR, and obtained the testing accuracy for other methods from their corresponding papers.

\textbf{Training:} Owing to the excellent pre-trained weights of the DINO-Mix backbone, most of the training weights were frozen during the training process. However, to make it more suitable for the VPR task, we fine-tuned the end of the backbone and trained the feature aggregation module. We trained DINO-Mix following the standard framework proposed in GSV-Cities~\cite{ali-bey_gsv-cities_2022}, which introduces a high-precision dataset consisting of 67k locations described by 560 k images. The batch size B was flexibly adjusted based on the model parameter size, and each location was trained with four images, resulting in a mini-batch of B × 4 images. Stochastic gradient descent~\cite{ruder_overview_2017} (SGD) with a momentum of 0.9 and a weight decay of 0.001 was employed for optimization. The initial learning rate was set to 0.05 and was divided by three every five epochs. Finally, the model was trained using images resized to 224 × 224 pixels over 50 epochs. Most existing VPR studies employ a triplet loss function based on weak supervision~\cite{hermans_defense_2018} for network training; however, this approach requires significant GPU memory and has a high computational overhead. Thus, we utilized multi-similarity loss~\cite{wang_multi-similarity_2019} as the training loss function. Multi-similarity loss mitigates the issues of large interclass distances and small intraclass distances in metric learning by considering multiple similarities. Instead of relying solely on absolute spatial distance as the sole metric, it uses the overall distance distribution of other sample pairs within a batch size to weigh the loss. This computational approach effectively promotes model convergence in the early stages, as expressed by Equation \ref{equation7}:

\begin{equation}
\label{equation7}
\begin{split}
    L_{MS} = \frac{1}{m}\sum_{i=1}^{m}  \bigg\{\frac{1}{\alpha}  { \log \big[1 + \sum_{k  \in P_{i} } e^{-\alpha (S_{ik} - \lambda)}}\big]  \\
    + \frac{1}{\beta }  { \log \big[1+ \sum_{k \in N_{i}}e^{\beta (S_{ik} - \lambda)} \big]} \bigg\}.
\end{split} 
\end{equation}

where $P_{i}$ represents the set of positive sample pairs for each instance in a batch size,  $N_{i}$ denotes the set of negative sample pairs for each instance in the same batch size, $S_{ij}$ and $S_{ik}$ represent the similarities between the two images, and $\alpha$, $\beta$, and $\lambda$ are hyperparameters.

\textbf{Evaluation:} In this work, we employed top-k accuracy~\cite{huang_survey_2023} as a metric to evaluate the precision of the VPR methods. Top-k accuracy is a commonly used evaluation method in the VPR domain, where it is considered successful if at least one of the top-k localization results for a query image has a geographical distance of less than a threshold s from the true location. In our experiments, we set s to 25 $m$ to align with existing methods.

\subsection{Comparison to the State-Of-The-Art}
\label{Comparison to the state-of-the-art}

Based on the conclusions drawn from the ablation studies in this work, we adopt the ViTb14 pre-trained model, which exhibits the best performance among the four models of DINOv2 as the backbone network for DINO-Mix in the VPR task, and modify the DINOv2 model by removing its Layer Norm and Head modules. We utilize Mix as the feature aggregation module to construct the model. During training, we update the parameters of the last three blocks of the backbone network and the entire Mix feature aggregation module. The number of Feature Mixer blocks in the Mix feature aggregation module is set to 2, and the dimensionality of the image features output by the model is 4096. By utilizing these optimal parameter settings, we conduct tests on six test sets for DINO-Mix and compare it with existing methods, as shown in Tab.\ref{tab:Test Results of Different Methods}. In addition, this paper presents Fig.\ref{fig:teaser} to more visually demonstrate the accuracy difference between DINO-Mix and other major VPR methods.

We adopted the ViTb14 pre-trained model, which exhibited the best performance among the four models of DINOv2, as the backbone network for DINO-Mix in the VPR task and modified the DINOv2 model by removing its layer norm and head modules. We used Mixer as a feature aggregation module to construct the model. During training, we updated the parameters of the last three blocks of the backbone network and the entire mix feature aggregation module. The number of feature mixer blocks in the mix feature aggregation module was set to two, and the dimensionality of the image features output by the model was 4096. By utilizing these optimal parameter settings, we conducted tests on six test sets for DINO-Mix and compared them with the existing methods, as shown in Tab.\ref{tab:Test Results of Different Methods}. In addition, Fig.\ref{fig:teaser} illustrates the difference in accuracy between DINO-Mix and other major VPR methods.

\begin{table*}[h]
\renewcommand{\thetable}{3}
    \caption{\emph{\textbf{Table of Test Results of Different Methods on Datasets with Changes in Viewpoint, Illumination, Season.} DINO-Mix(ViTb14) is ours, Bolded numbers are optimal results, and underlined numbers are sub-optimal results}}
    \centering
    \begin{tabular}{c p{1.6cm} c p{0.9cm} c p{0.8cm}  c p{0.9cm}  c p{0.8cm}  c p{0.8cm} c p{0.8cm}  c p{0.8cm}  c p{0.8cm}}
    \hline
    Method & Training data & Vector dim & & & & Test dataset\\
    
    &&& Pitts250k & Pitts30k & SF-XLval & Tokyo24/7 & Nordland & SF-XLTestv1 \\
    \hline
    MAX~\cite{relja_netvlad_2018}   & GSV-cities& 1024& 46.45& 55.87 &43.68&3.17&9.30&8.10 \\
    AVG~\cite{relja_netvlad_2018} & GSV-cities& 1024 &51.85& 64.20 &45.43&8.25&14.99&13.90 \\
    SPOC~\cite{yandex_aggregating_2015}  & GSV-cities & 256 & 60.59 & 68.37 &56.11&19.68&10.18&20.50 \\
    MAC~\cite{razavian_visual_2016}  & GSV-cities  & 256 & 61.75& 69.42 &56.47&18.73&13.49&22.00 \\
    RMAC~\cite{tolias_particular_2016} & GSV-cities  & 256 & 71.3  & 75.94&63.74&32.70&13.30&29.40 \\
    RRM~\cite{kordopatis-zilos_leveraging_2021}   & GSV-cities  & 256 &88.14& 87.49&81.60&57.46&46.00&53.40 \\
    GeM~\cite{radenovic_fine-tuning_2019}  & GSV-cities  & 256  &76.01 & 79.61&68.79&34.60&23.8&33.70 \\
    NetVLAD~\cite{relja_netvlad_2018}& Pitts-30k  & 16384 & 86.93& 86.36 &65.34&53.97&7.86&42.50 \\
    NetVLAD~\cite{relja_netvlad_2018} & GSV-cities& 16384& 89.71& 88.04 &80.38&70.79&36.25&58.90 \\
    CRN~\cite{kim_learned_2017}  & GSV-cities &  16384& 90.60& 89.03&81.83&70.16&38.58&64.40 \\
    MultiRes-NetVLAD~\cite{khaliq_multires-netvlad_2022} &Pitts-30k&32768& 86.70& 86.80 &--&69.80&--&-- \\
    SARE~\cite{liu_stochastic_2019} & Pitts-30k& 4096 & 88.00&87.20&--&74.80&--&45.5 \\
    SERS~\cite{ge_self-supervising_2020} & Pitts-30k& 4096 &90.40 & 89.10 &--&80.30&16.00&50.30 \\
    CosPlace~\cite{berton_rethinking_2022}  & GSV-cities& 4096 & 89.89 & 88.54&84.01&63.17&47.62&54.30 \\
    ConvAP~\cite{ali-bey_gsv-cities_2022}  & GSV-cities& 4096 & 91.52& 89.67 &76.95&72.06&63.93&59.20 \\
    MixVPR~\cite{ali-bey_mixvpr_2023}(Resnet18)&GSV-cities& 4096& 91.75& 89.57&80.68&75.24&64.75&64.90 \\
    MixVPR~\cite{ali-bey_mixvpr_2023}(Resnet50)&GSV-cities& 4096& \underline{94.13} & \underline{91.52}&\underline{85.40}&\underline{85.40}&\underline{76.12}&\underline{75.70} \\
\textbf{DINO-Mix(ViTb14)(Ours)}&GSV-cities& 4096&\textbf{94.58} &\textbf{92.03}&\textbf{89.25}&\textbf{91.75}&\textbf{80.18}&\textbf{82.00} \\
    \hline
    \end{tabular}
    \label{tab:Test Results of Different Methods}
\end{table*}

As listed in Tab.\ref{tab:Test Results of Different Methods}, the test accuracy of the DINO-Mix model proposed in this paper has comprehensively surpassed that of the SOTA method, with further improvement in the Pittsburgh250k, Pittsburgh30k, and SF-XL-Val test sets focusing on changes in viewpoints, and especially in the Tokyo24/7, Nordland, and SF-XL-Testv1 test sets with changes in complex appearance environments.

\subsection{ablation studies}
\label{ablation studies}
    \subsubsection{Hyperparameters}

\begin{figure}[!t]
    \renewcommand{\thefigure}{5} 
        \centering
        \includegraphics[width=1.0\linewidth]{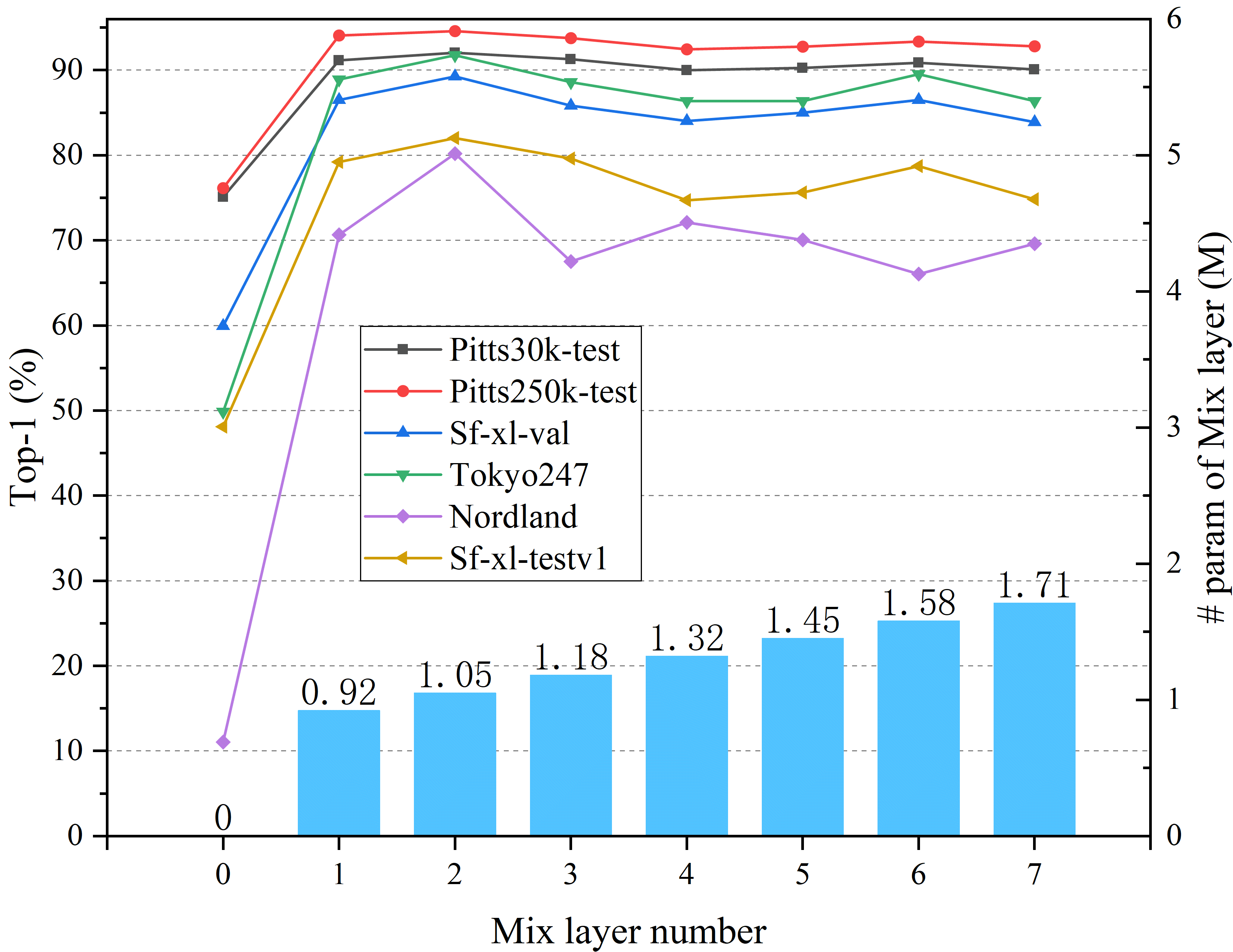}
        \vspace{-1em}
        \caption{\emph{\textbf{Ablation on the number of Feature-Mixer blocks}}}
        \label{fig:mix_layer_ablatio}
    \end{figure}

In DINO-Mix, the number of layers $L$ in the feature mixer is also a critical factor for image retrieval accuracy. To determine the optimal number of Mixer layers, we conducted tests on the Pitts30k-test Pitts250k-test, Sf-xl-val, Tokyo24/7, Nordland, and Sf-xl-testv1 datasets with different numbers of mix layers $L$ (1, 2, 3, 4, 5, 6, 7) for DINO-Mix using ViTb14 as the backbone network. The TOP-1 accuracy is depicted in Fig.\ref{fig:mix_layer_ablatio}. A careful examination of the figure reveals that without any Mix layers, DINO-Mix exhibits a lackluster test accuracy across all six datasets. However, upon incorporating one Mix layer, there is a remarkable enhancement in test accuracy. This observation highlights the pivotal role played by the feature aggregation module in elevating the precision of DINO-Mix. As the number of Mix layers further increases up to two, there is a marginal improvement in test accuracy, culminating in a peak. Nevertheless, as the number of Mix layers continues to escalate, DINO-Mix's test accuracy on the six datasets displays a slow decline and fluctuations, accompanied by a linear increase in parameters. Based on the above analysis, this study adopts a two-layer Mix scheme as the feature aggregator in DINO-Mix.

\subsubsection{Descriptor dimensionality}

\begin{figure}[!t]
    \renewcommand{\thefigure}{6} 
        \centering
        \includegraphics[width=1.0\linewidth]{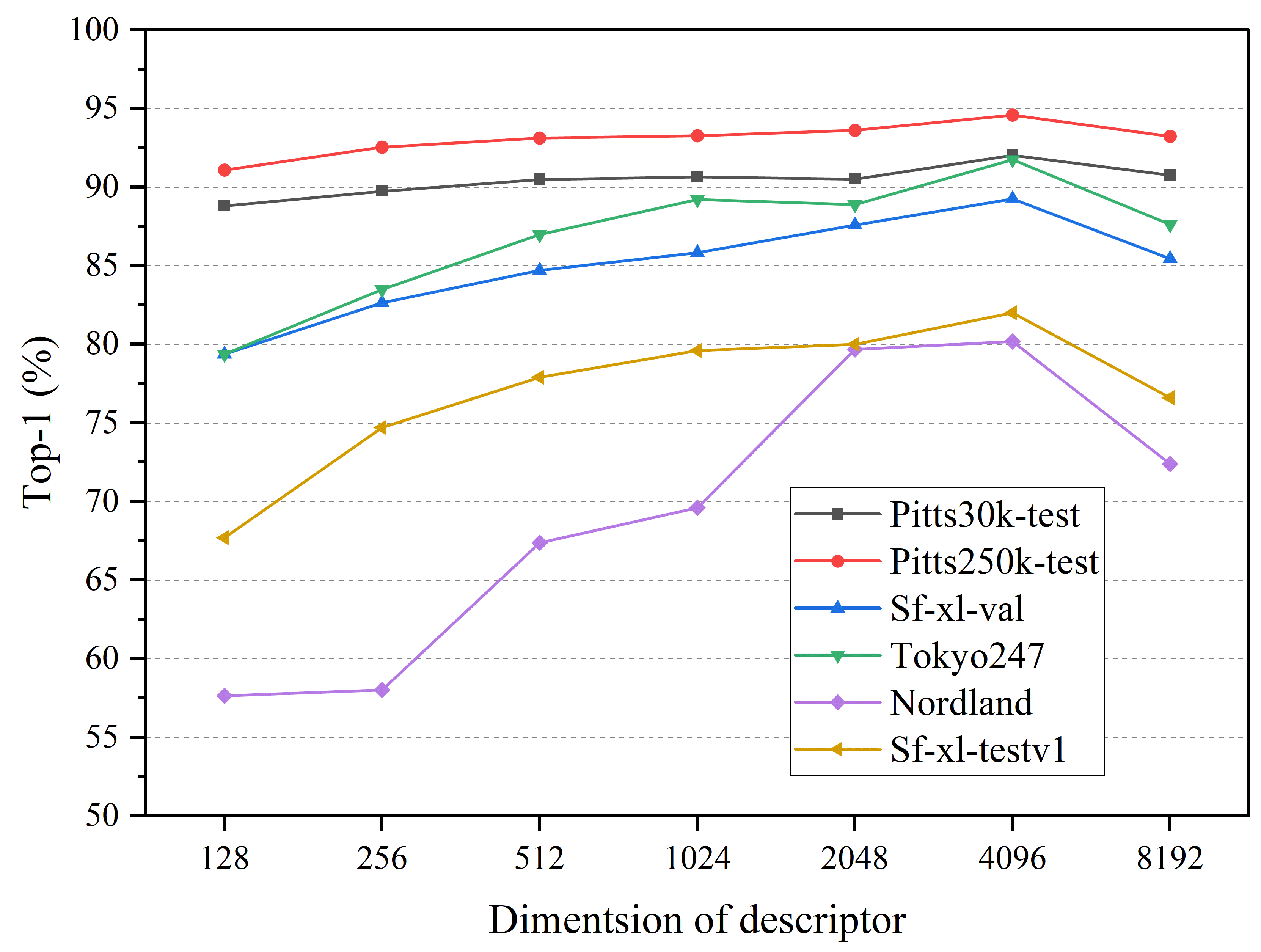}
        \vspace{-1em}
        \caption{\emph{\textbf{performance on potts30k-test with different dimensionality configurations.}}}
        \label{fig:dim_ablation}
    \end{figure}

We conducted an ablation study on the dimensionality of image feature vectors extracted using the DINO-Mix. The experiment employed ViTb14, which exhibited the best performance as the backbone network, with two layers in the Mixer module. The test was used as the Pitts30k-test, Pitts250k-test, Sf-xl-val, Tokyo24/7, Nordland, and Sf-xl-testv1 datasets, and the image feature vector dimensionality was varied by changing the number of channels in the output vector of the Mixer module. The tested dimensions of the image feature vectors were 128, 256, 512, 1024, 2048, 4096, and 8192. As depicted in Fig.\ref{fig:dim_ablation}, an increase in the dimensionality of image feature vectors is observed to have a positive impact on the overall Top-1 test accuracy of DINO-Mix across various datasets. This trend is particularly pronounced in Sf-xl-val, Tokyo247, Nordland, and Sf-xl-testv1 datasets, where there is a rapid rise in accuracy. Ultimately, the highest level of accuracy is achieved at a dimensionality of 4096. This phenomenon suggests that utilizing image feature vectors with too low dimensionality may result in reduced robustness to variations such as changes in illumination and seasonal shifts in VPR tasks. Consequently, this study adopts a final image feature dimensionality of 4096.

   \subsubsection{Backbone architecture} 

\begin{figure}[!t]
    \renewcommand{\thefigure}{7} 
        \centering
        \includegraphics[width=1.0\linewidth]{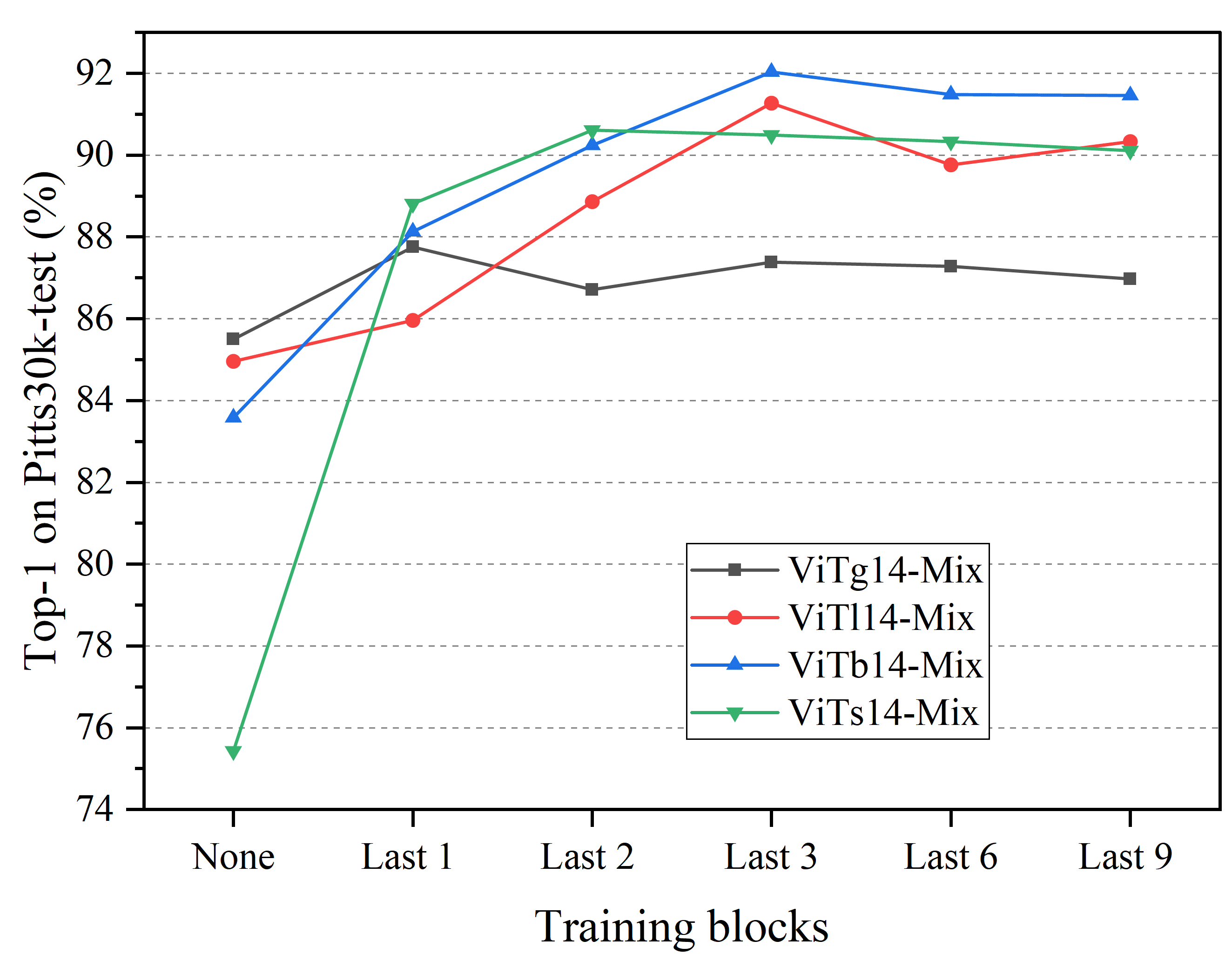}
        \vspace{-1em}
        \caption{\emph{\textbf{Test results of different DINO-Mix models with different weights for updating the number of layers.}}}
        \label{fig:backbone_ablation}
    \end{figure}

DINOv2 encompasses four ViT models, with ViTg14 (giant) being the largest. Through model knowledge distillation, three smaller models were obtained from the distillation process, including ViTl14 (large), ViTb14 (base), and ViTs14 (small), as displayed in Tab.\ref{tab:Four ViT model param}. To evaluate the performance of these four models in the DINO-Mix framework, we conducted training on GSV-Cities as the training set and tested Pitts30k-test using ViTg14-Mix, ViTl14-Mix, ViTb14-Mix, and ViTs14-Mix. The feature mixer was fixed at two layers and the dimensionality of the image feature vectors was set to 4096. In addition, we trained and tested the DINO-Mix models with four different backbone networks under six scenarios: updating the weights of the last one, two, three, six, and nine blocks, and not updating the weights of the backbone network (none). The results are shown in Fig.\ref{fig:backbone_ablation}.

From the perspective of the four differently sized backbone networks, ViTb14-Mix exhibited higher accuracy than the other three models, with a maximum Top-1 accuracy of $92.03\%$. In contrast, ViTg14-Mix exhibited the worst overall performance. This suggests that ViTg14’s large parameter count extracts deeper features from images, which adversely affects subsequent feature aggregation in the feature mixer.

Models without parameter updates for the backbone network demonstrated poorer performance. As the number of updated blocks increased, both ViTb14-Mix and ViTl14-Mix showed a gradual improvement in test accuracy, reaching their highest values after updating the parameters of the last three blocks, and stabilizing thereafter. In contrast, ViTs14-Mix achieved the highest test accuracy and stability after updating the parameters of the last two blocks. However, for ViTg14-Mix, the block parameter updates did not significantly enhance accuracy. Starting from the last three blocks, the ViTg14-Mix test accuracy showed a downward trend. This indicates that excessively deep block parameter updates may extensively alter the original pre-trained parameters.

\begin{figure}[!t]
    \renewcommand{\thefigure}{8} 
        \centering
        \includegraphics[width=1.0\linewidth]{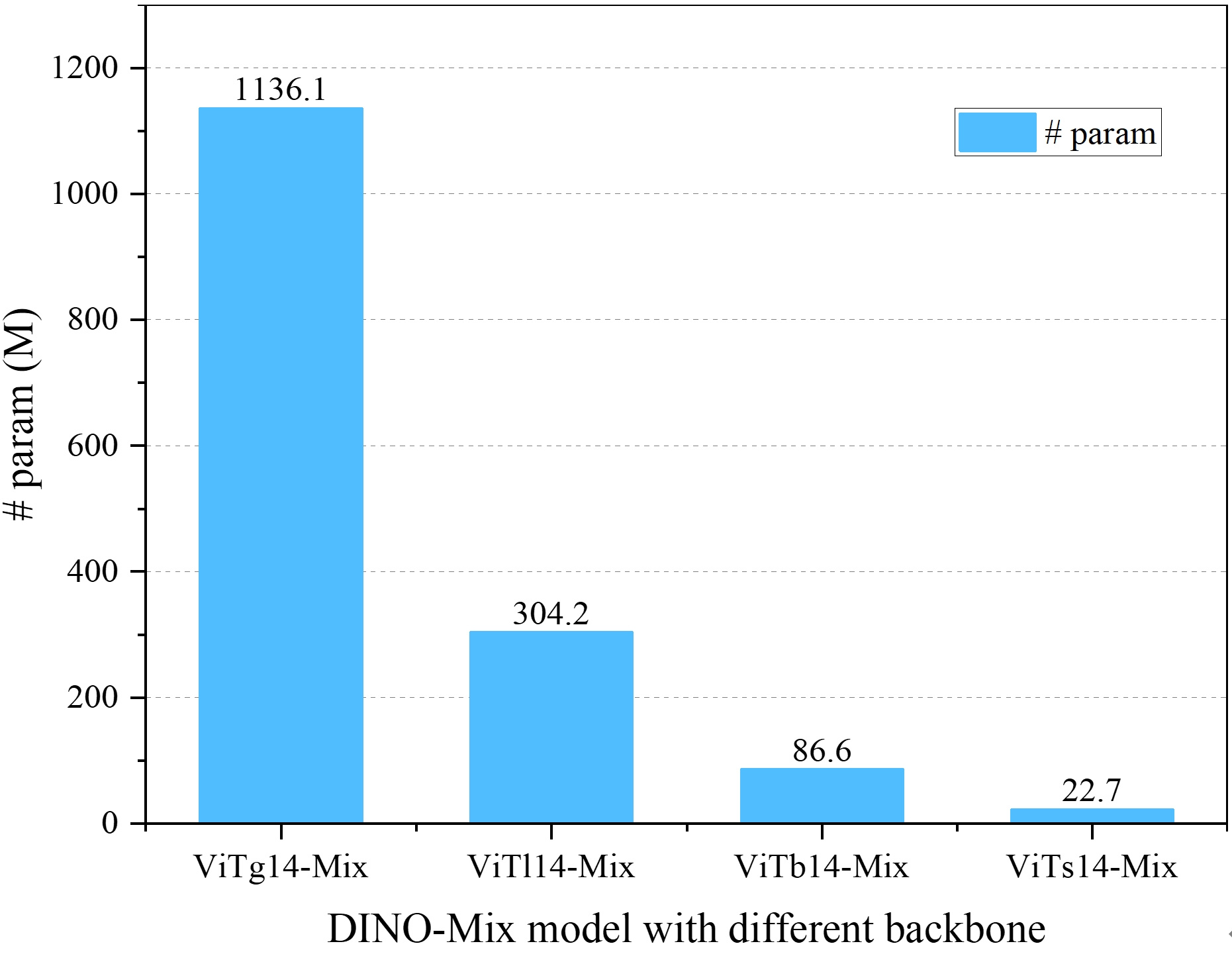}
        \vspace{-1em}
        \caption{\emph{\textbf{Parametric quantities of the four DINO-Mix models.}}}
        \label{fig:backbone_params}
    \end{figure}

In summary, updating the parameters of the last three blocks of the backbone network yielded optimal results. Considering the parameter counts of the four DINO-Mix models shown in Fig.\ref{fig:backbone_params}, we selected ViTb14-Mix, which has a moderate parameter count and superior test accuracy, as the final model for DINO-Mix.

%% file: Tex_content/qualitative_results.tex
\subsection{Image retrieval comparison} 
\label{image retrieval conparision}

            \begin{table*}[!t]
    \renewcommand{\thetable}{4}
    \caption{\emph{\textbf{Comparison of image retrieval results of DINO-Mix with other methods in difficult cases (Top-1).}The green and red boxes in the table represent image retrieval success and failure, respectively, and the yellow box represents that the image content should be correct but the localization distance exceeds the threshold s.}}
    \centering
    \begin{tabular}{c p{2cm} c p{2cm} c p{2cm} c p{2cm} c p{2cm} c p{2cm} c p{2cm}}
    \hline
    Category&Query&DINO-Mix&MixVPR&NetVLAD&ConvAP&CosPlace\\
    \hline
    Viewpoint Change&\includegraphics[width=0.1\textwidth]{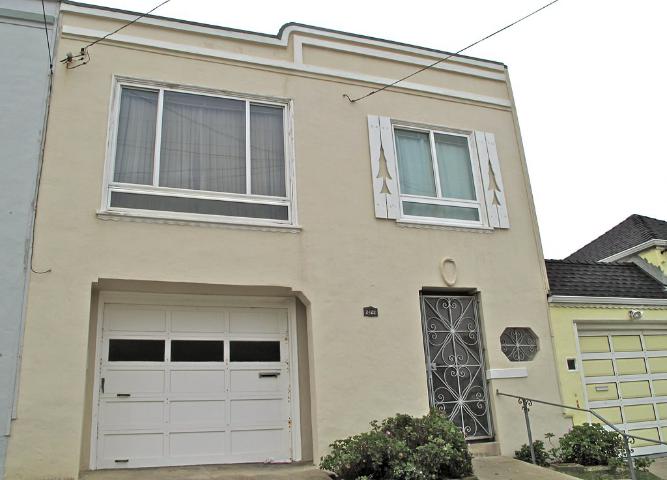}& \includegraphics[width=0.1\textwidth]{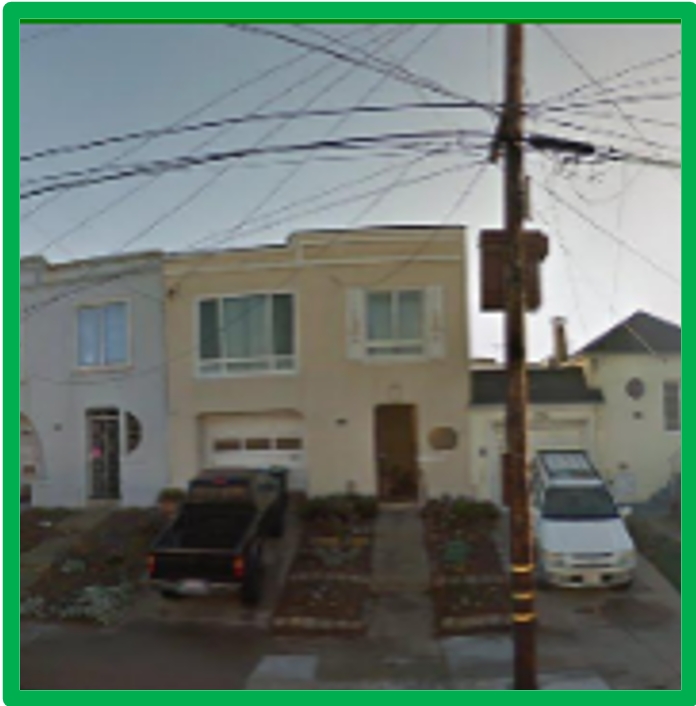}&\includegraphics[width=0.1\textwidth]{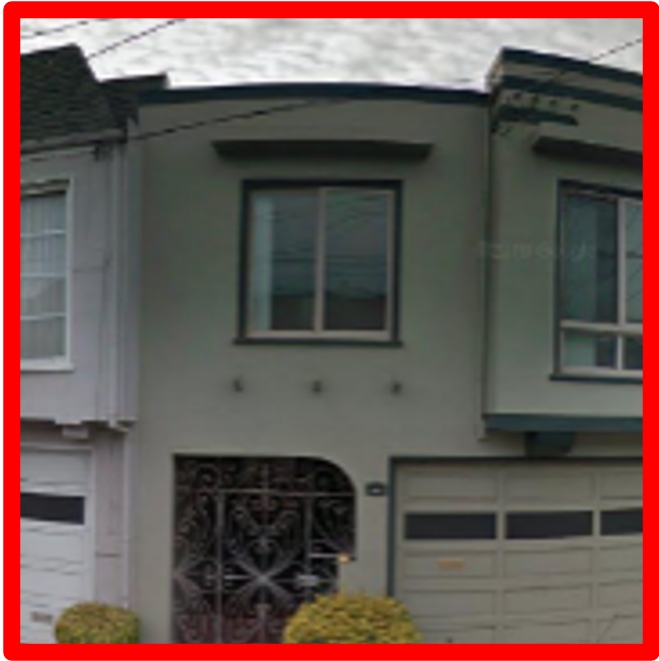}&\includegraphics[width=0.1\textwidth]{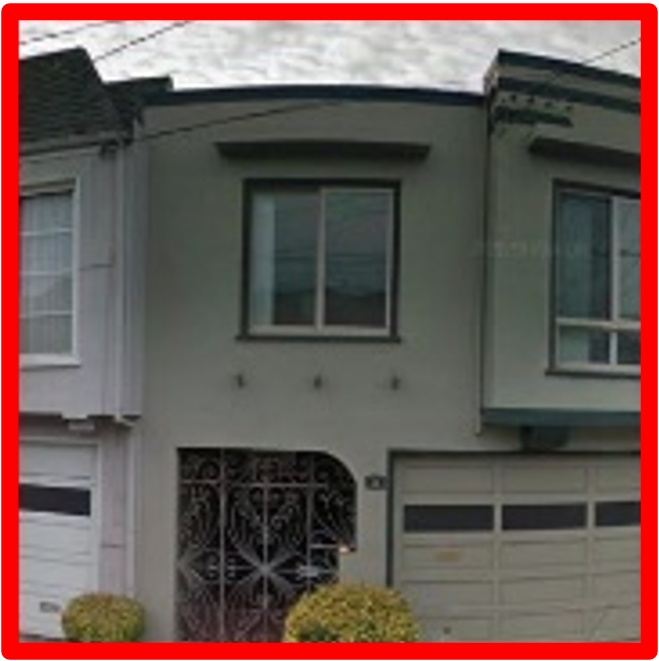}&\includegraphics[width=0.1\textwidth]{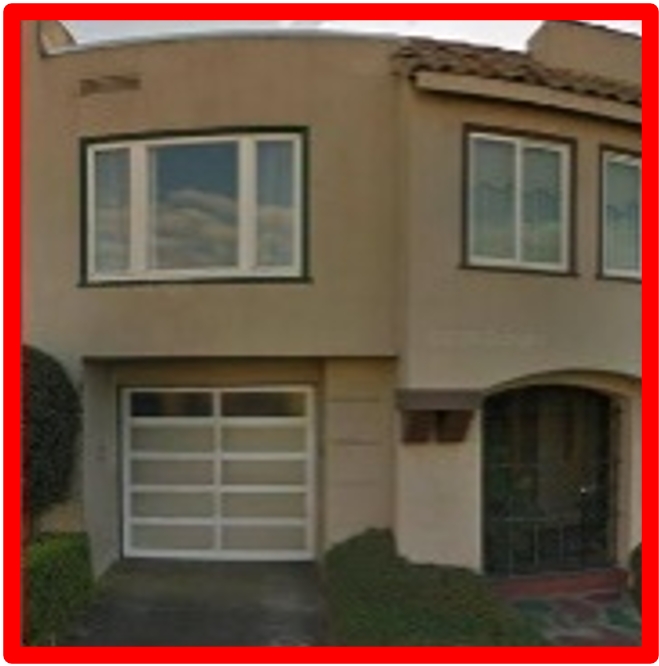}&\includegraphics[width=0.1\textwidth]{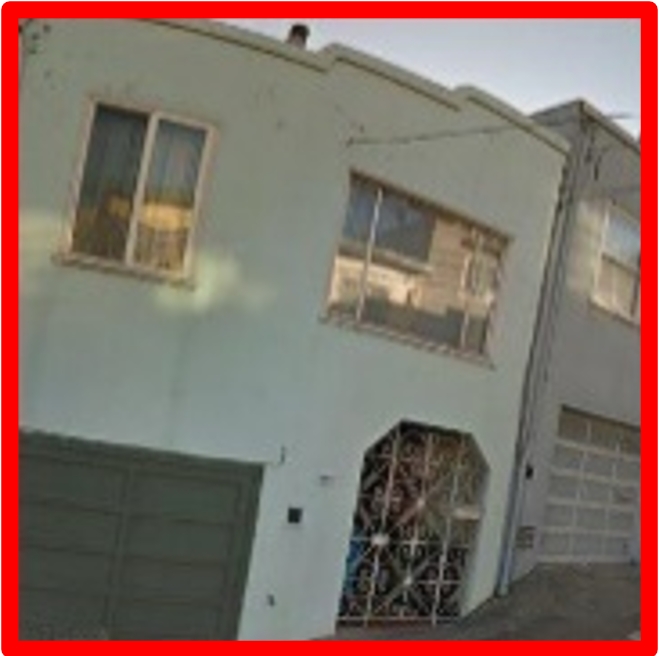} \\
     &\includegraphics[width=0.09\textwidth]{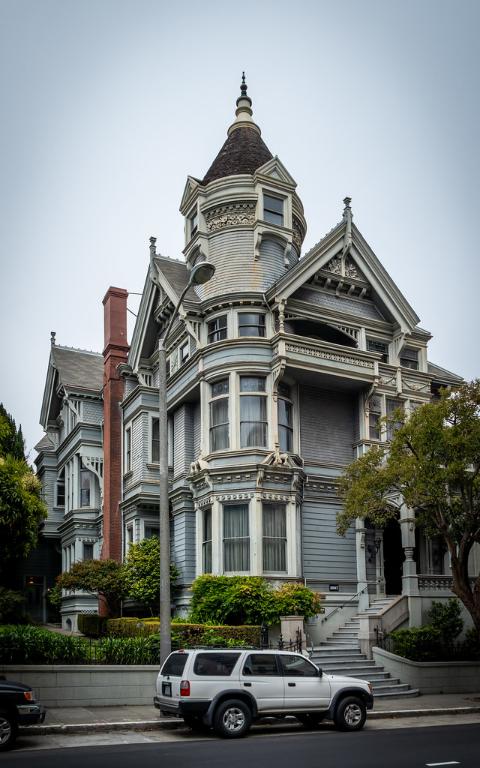}& \includegraphics[width=0.1\textwidth]{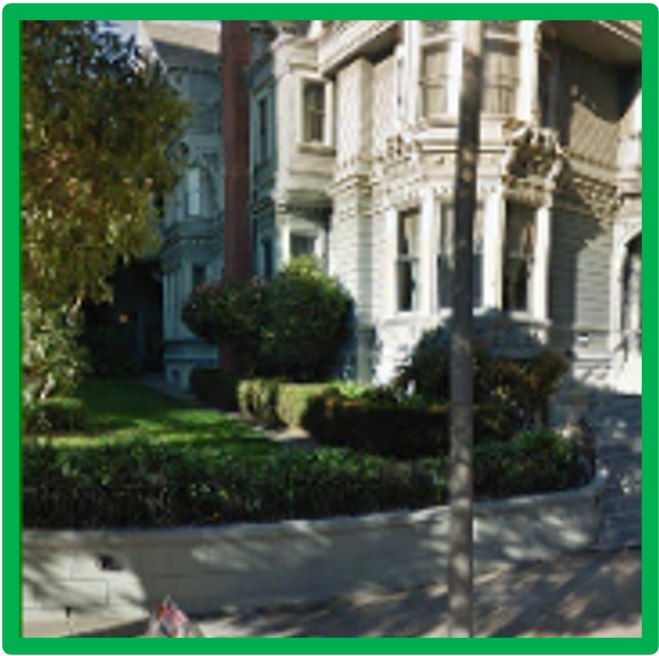}&\includegraphics[width=0.1\textwidth]{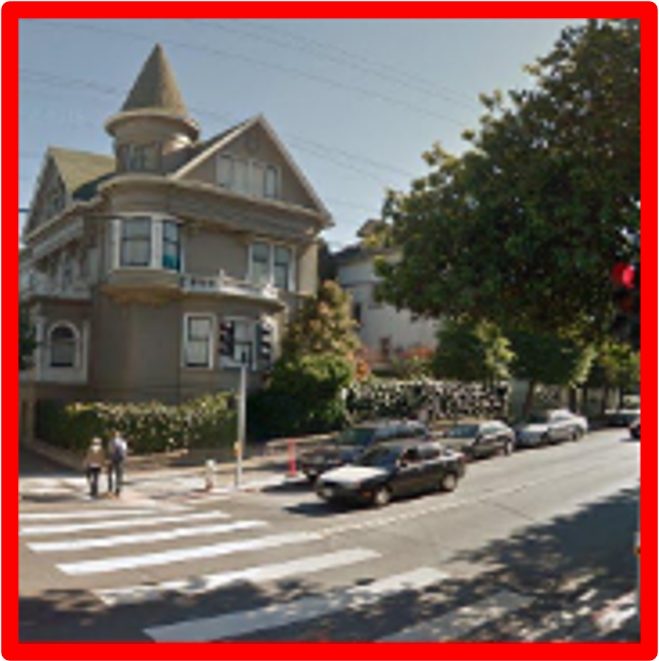}&\includegraphics[width=0.1\textwidth]{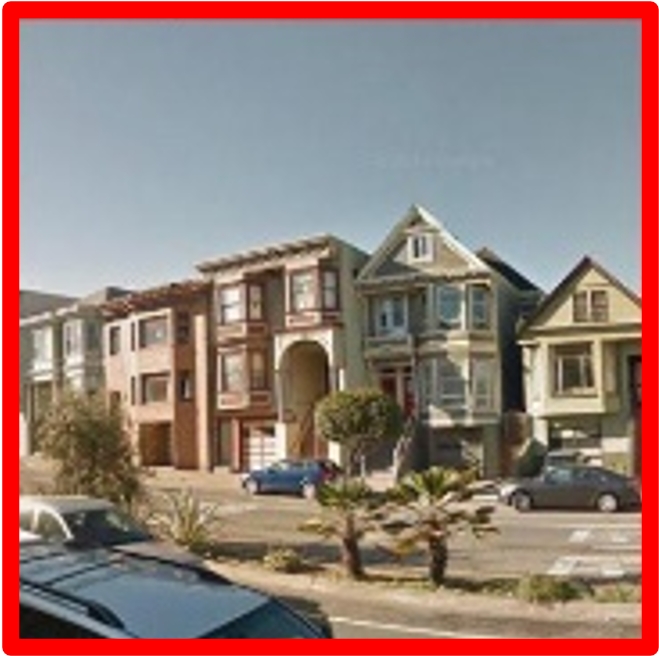}&\includegraphics[width=0.1\textwidth]{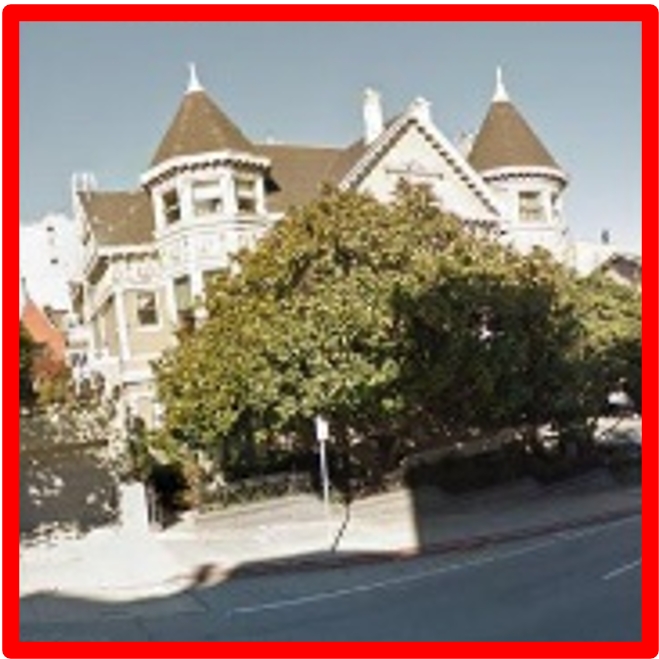}&\includegraphics[width=0.1\textwidth]{Pics/2-2.png} \\
     \hline
     &\includegraphics[width=0.07\textwidth]{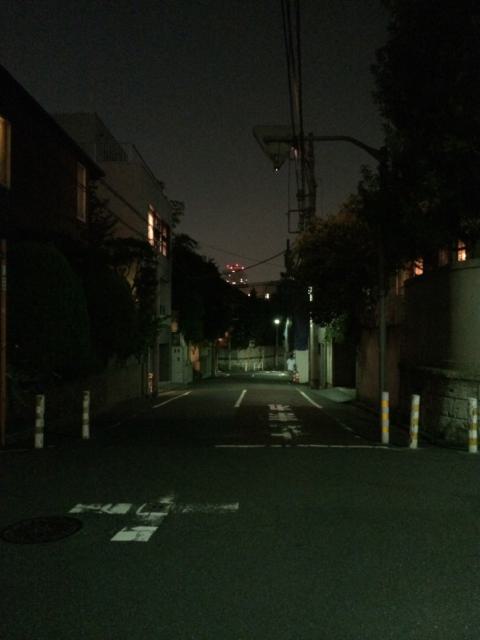}& \includegraphics[width=0.1\textwidth]{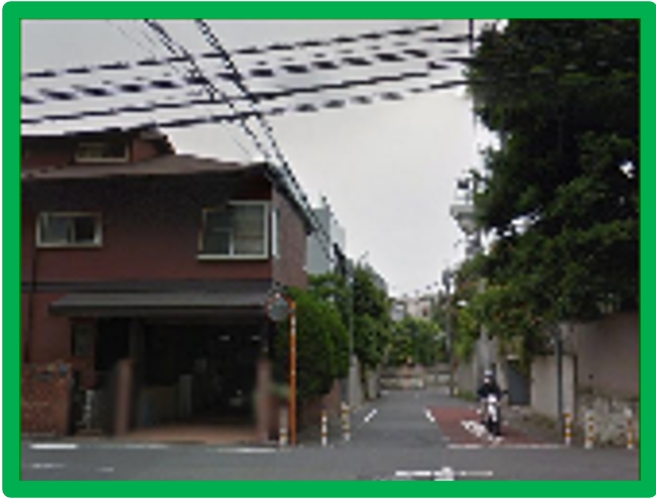}&\includegraphics[width=0.1\textwidth]{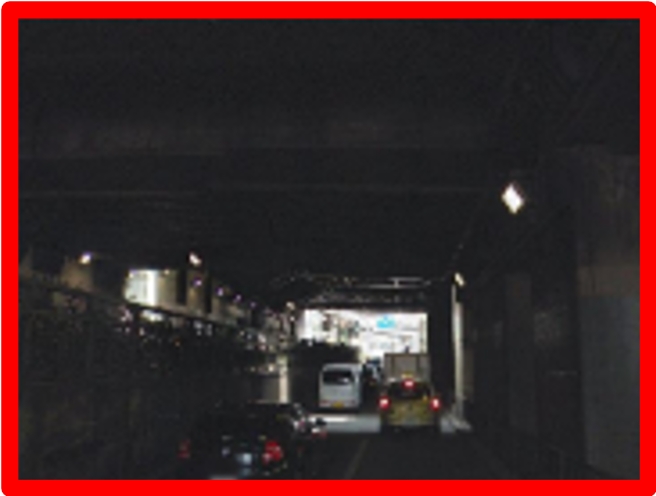}&\includegraphics[width=0.1\textwidth]{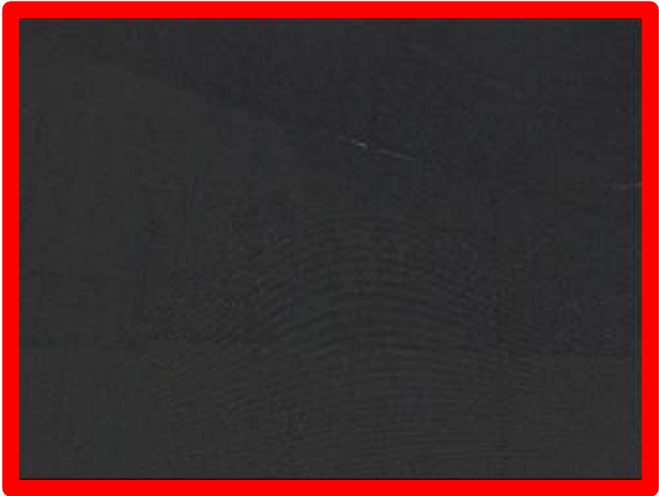}&\includegraphics[width=0.1\textwidth]{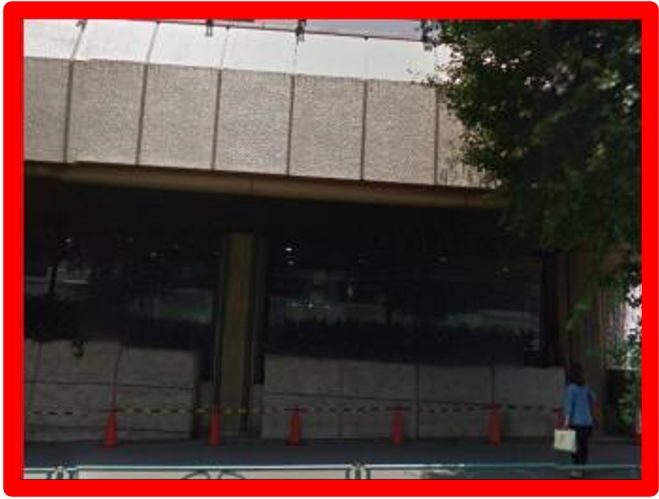}&\includegraphics[width=0.1\textwidth]{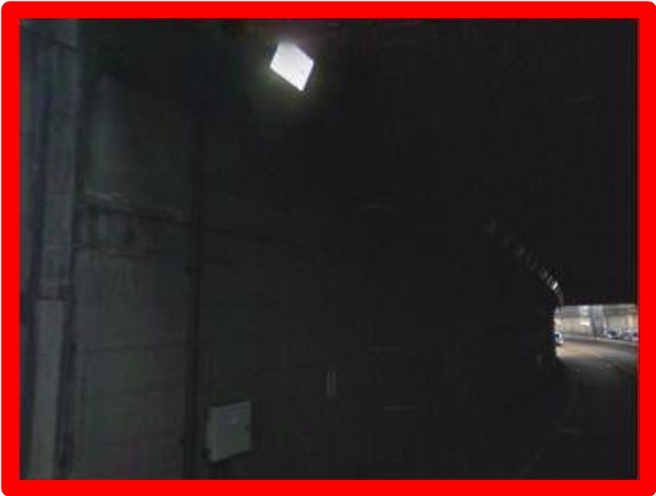} \\
    Illumination Change&\includegraphics[width=0.07\textwidth]{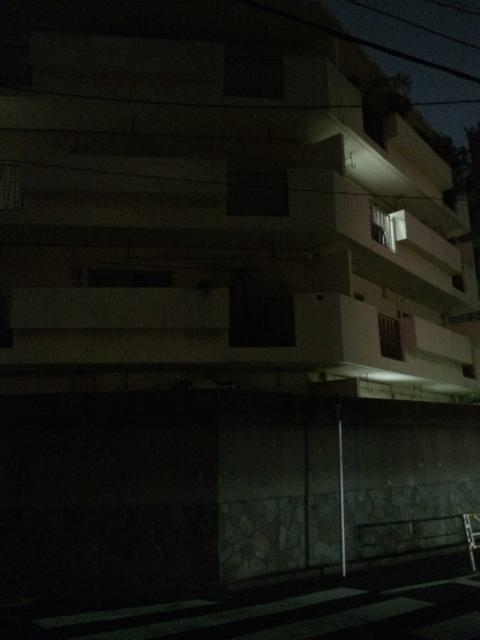}& \includegraphics[width=0.1\textwidth]{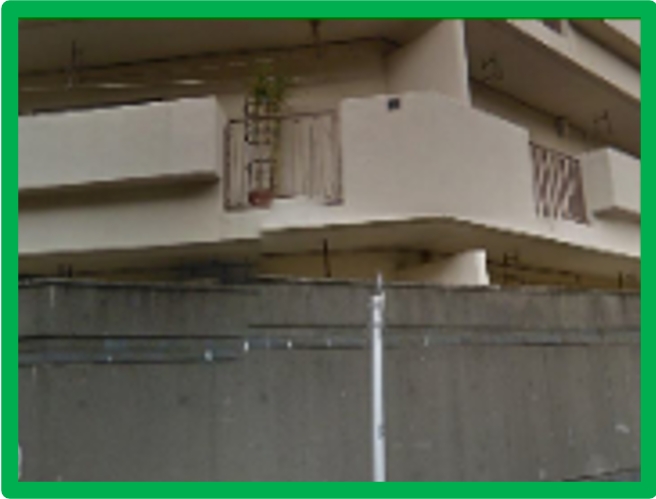}&\includegraphics[width=0.1\textwidth]{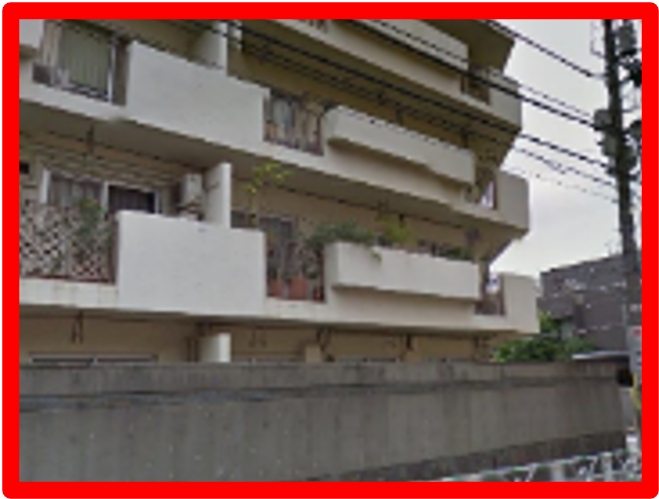}&\includegraphics[width=0.1\textwidth]{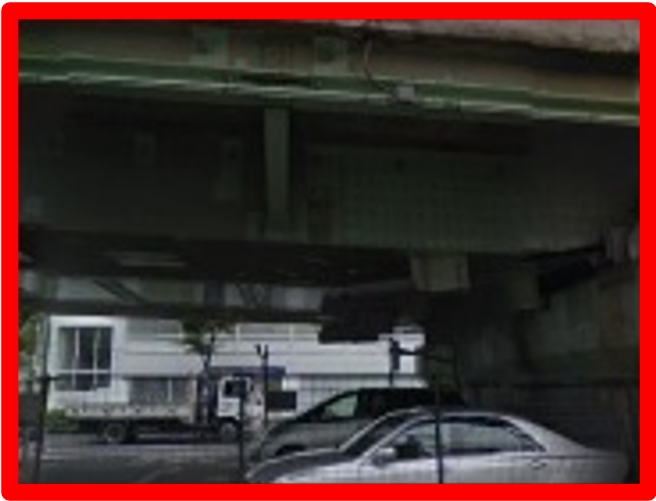}&\includegraphics[width=0.1\textwidth]{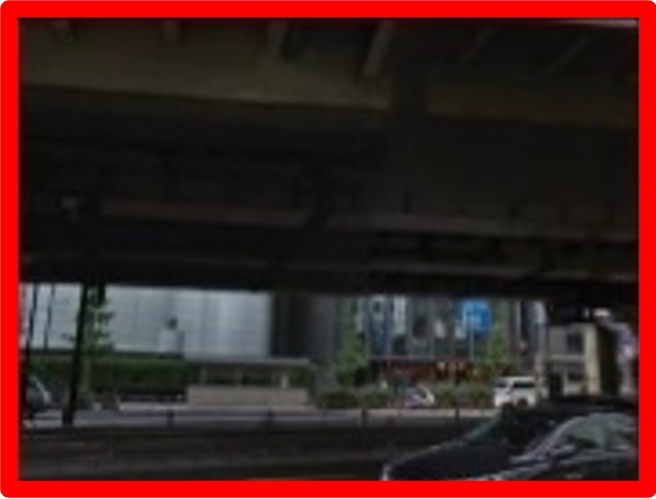}&\includegraphics[width=0.1\textwidth]{Pics/4-3.png} \\
    &\includegraphics[width=0.1\textwidth]{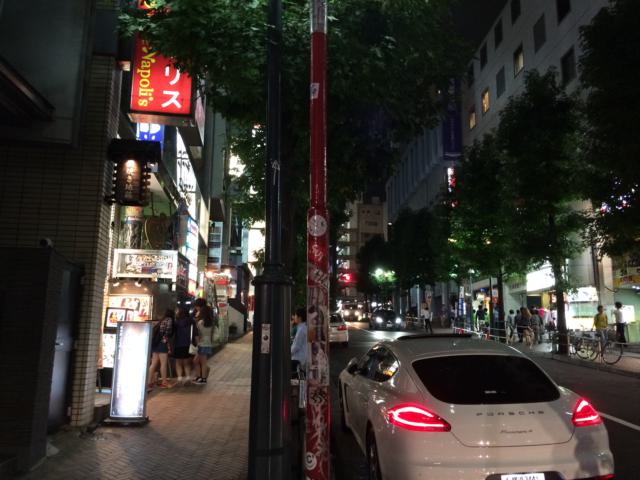}& \includegraphics[width=0.1\textwidth]{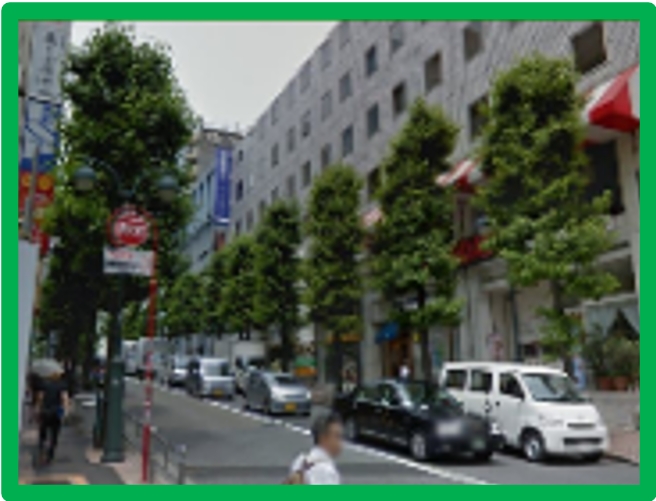}&\includegraphics[width=0.1\textwidth]{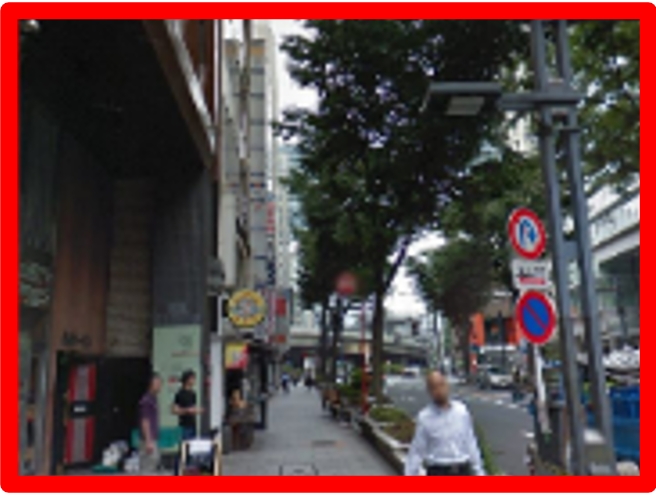}&\includegraphics[width=0.1\textwidth]{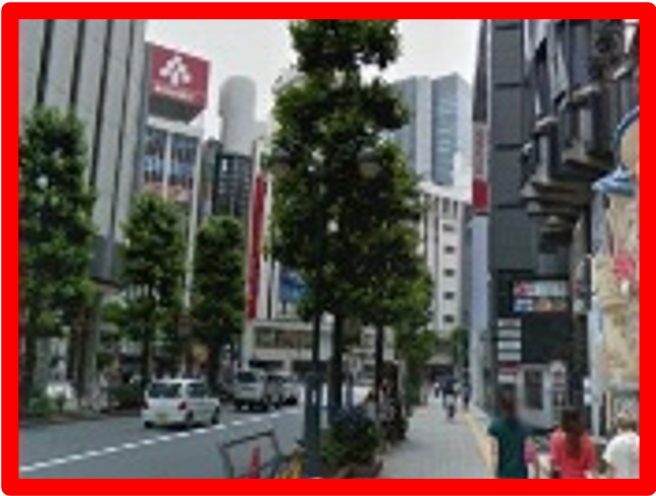}&\includegraphics[width=0.1\textwidth]{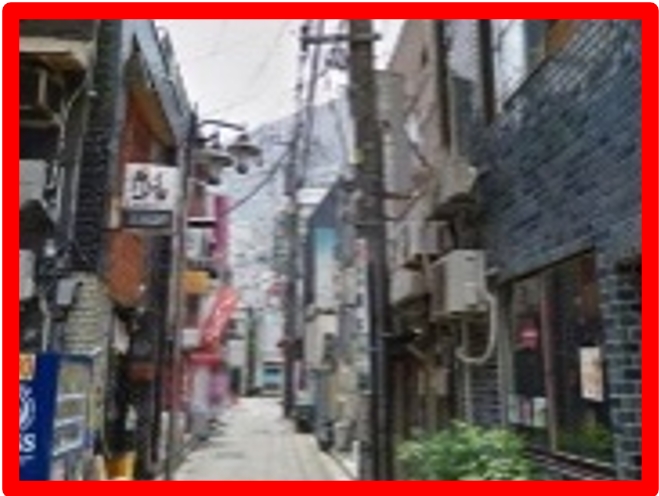}&\includegraphics[width=0.1\textwidth]{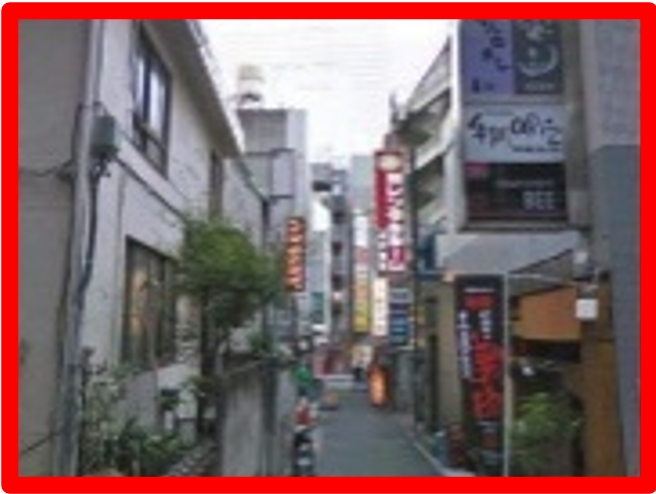} \\
    &\includegraphics[width=0.08\textwidth]{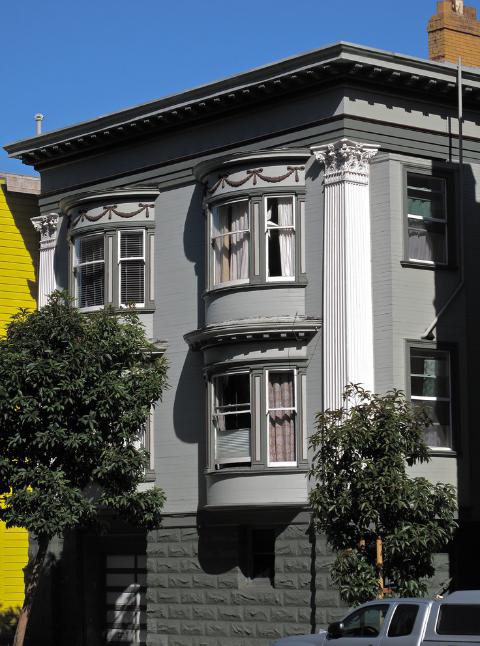}& \includegraphics[width=0.1\textwidth]{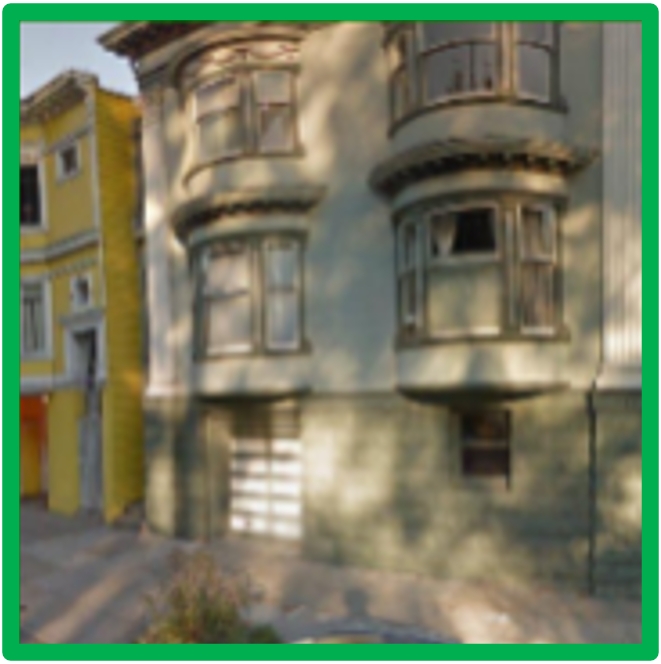}&\includegraphics[width=0.1\textwidth]{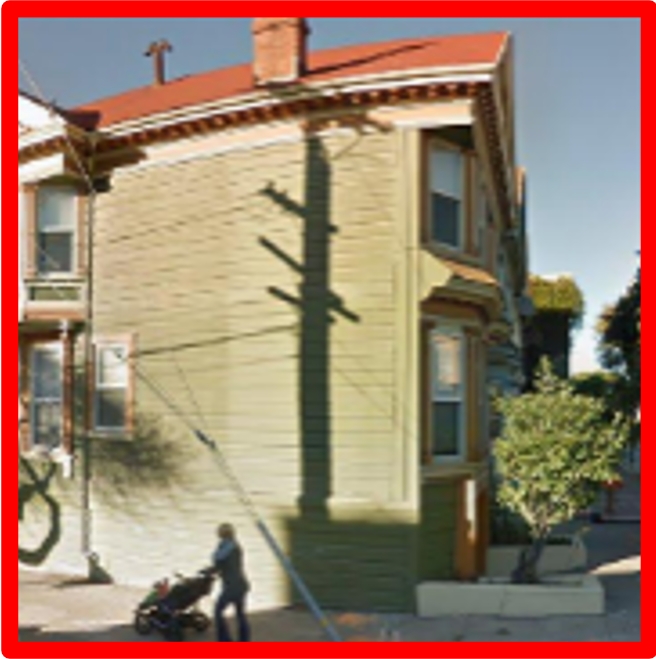}&\includegraphics[width=0.1\textwidth]{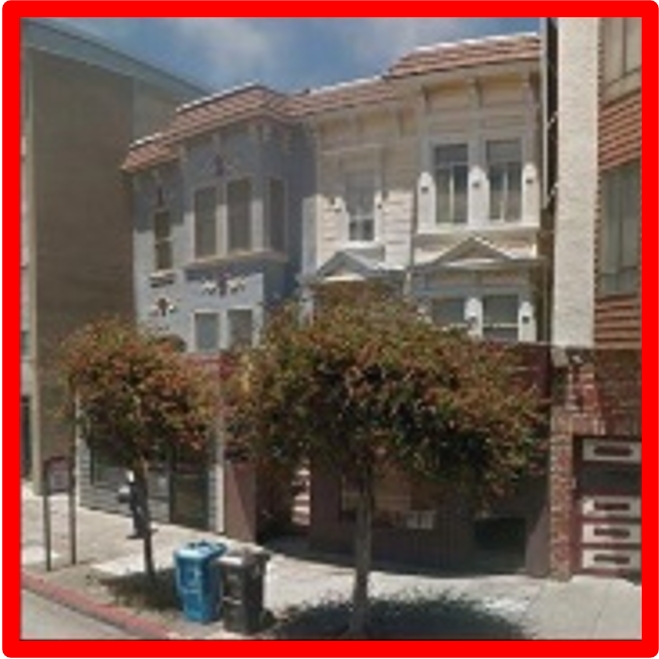}&\includegraphics[width=0.1\textwidth]{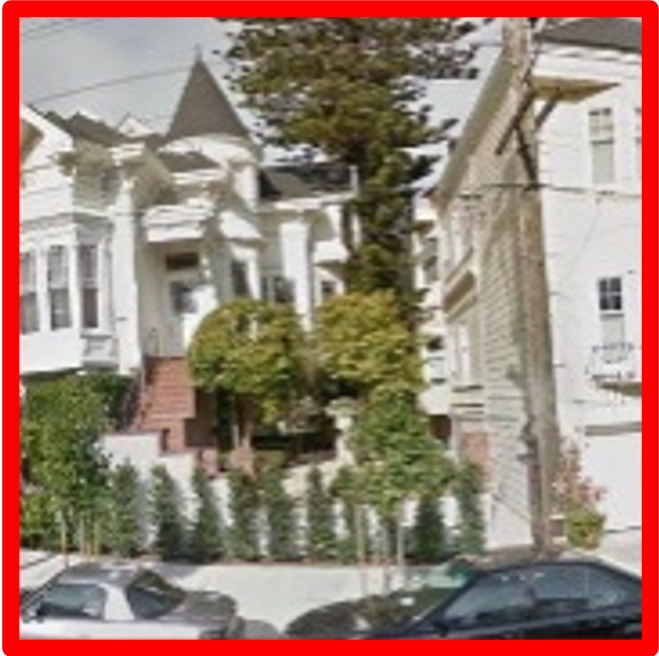}&\includegraphics[width=0.1\textwidth]{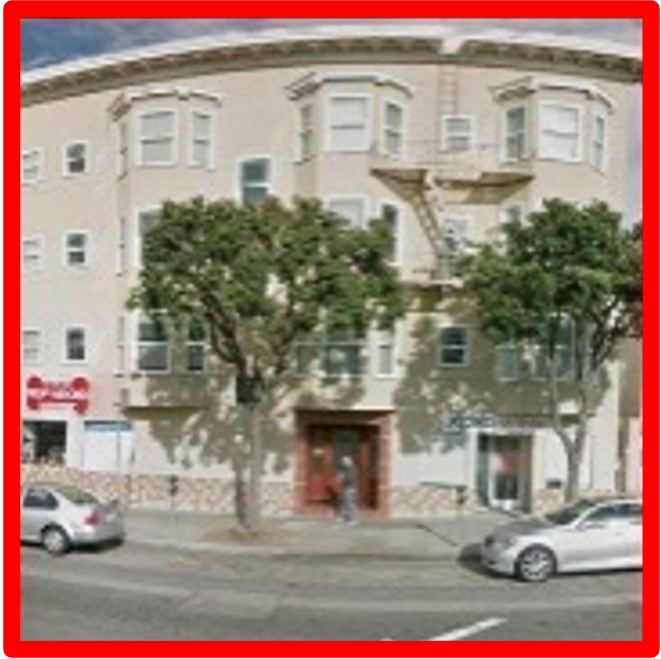} \\
    \hline
    Occlusions&\includegraphics[width=0.09\textwidth]{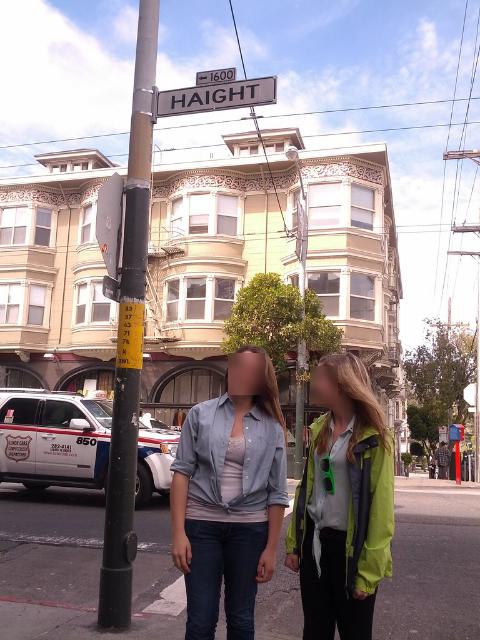}& \includegraphics[width=0.1\textwidth]{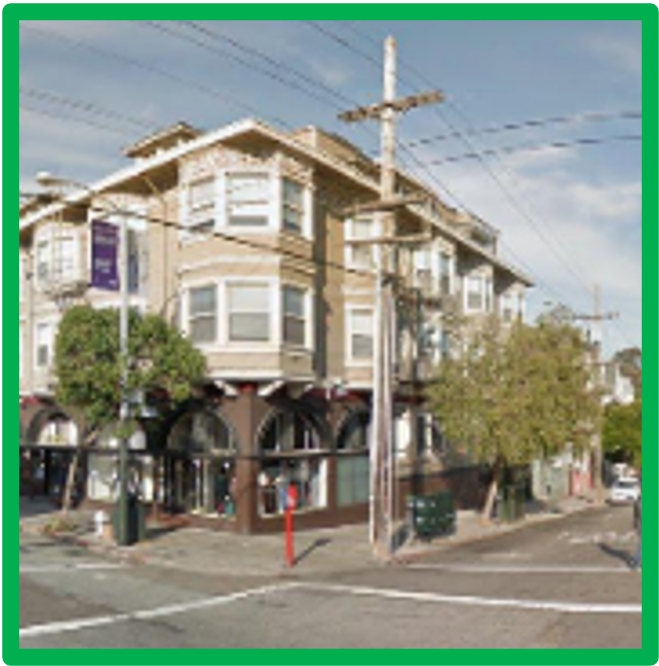}&\includegraphics[width=0.1\textwidth]{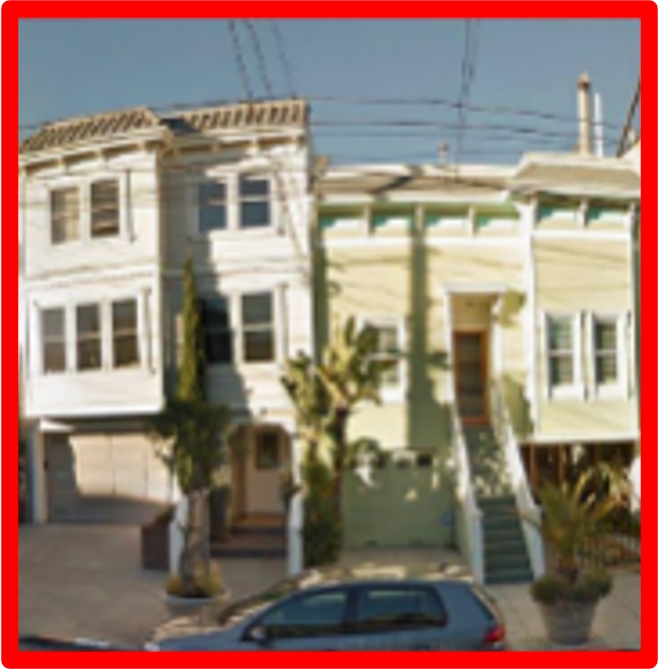}&\includegraphics[width=0.1\textwidth]{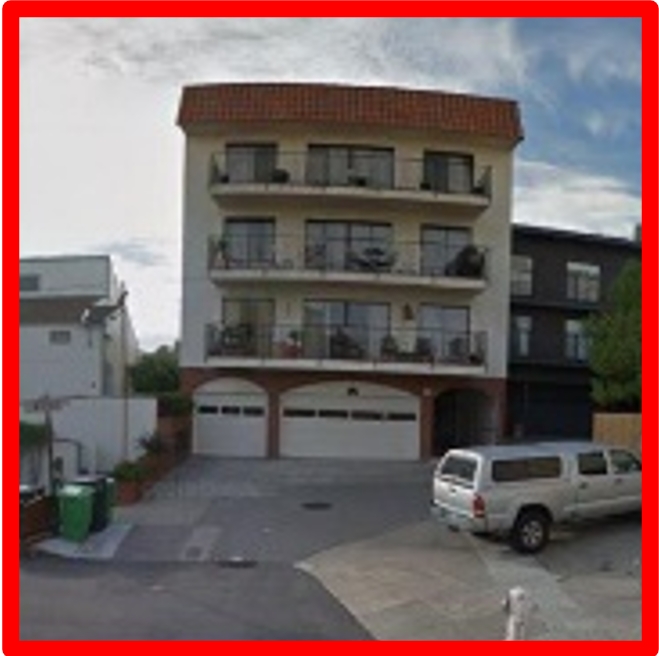}&\includegraphics[width=0.1\textwidth]{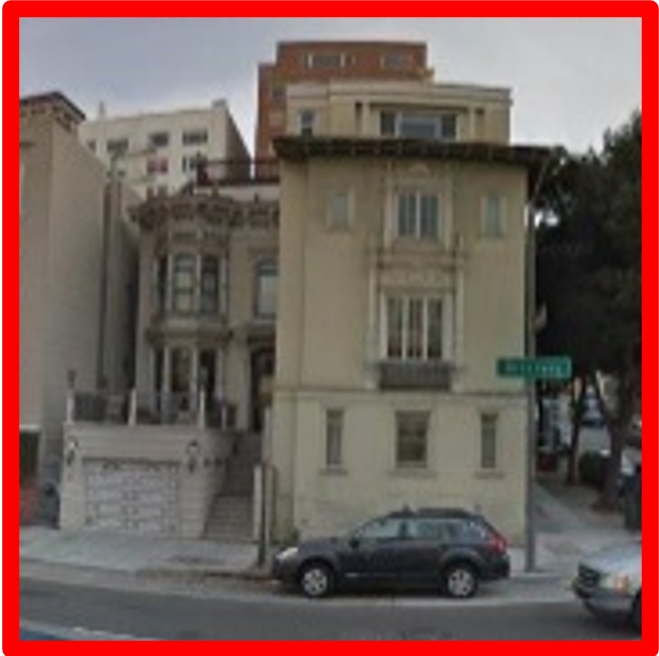}&\includegraphics[width=0.1\textwidth]{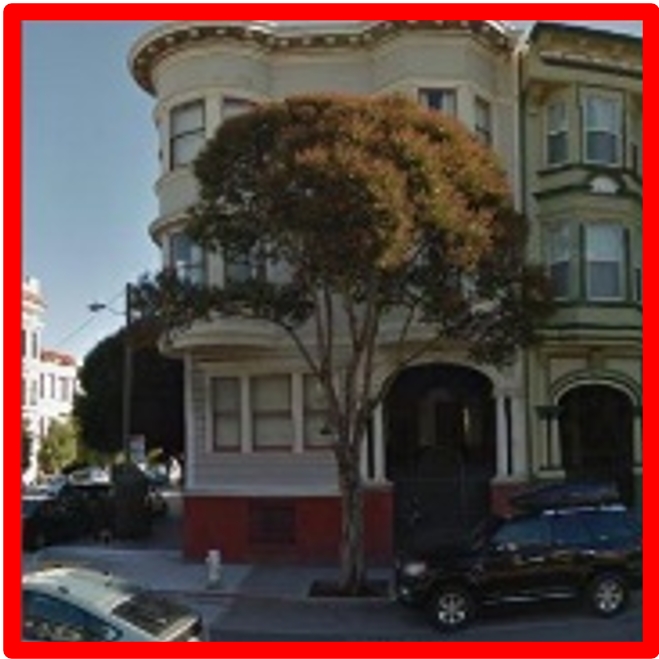} \\
    
   &\includegraphics[width=0.09\textwidth]{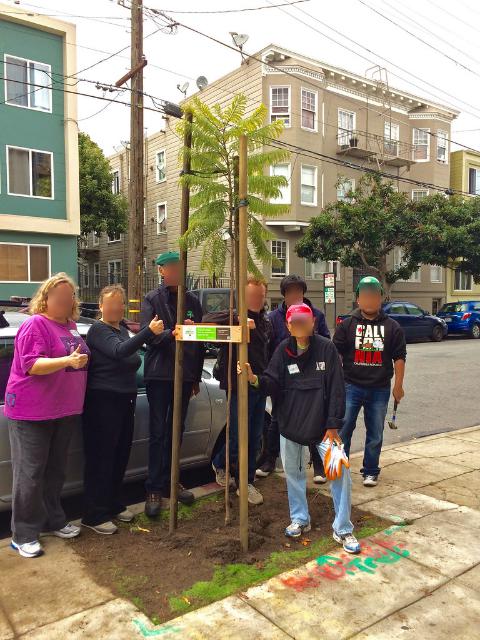}& \includegraphics[width=0.1\textwidth]{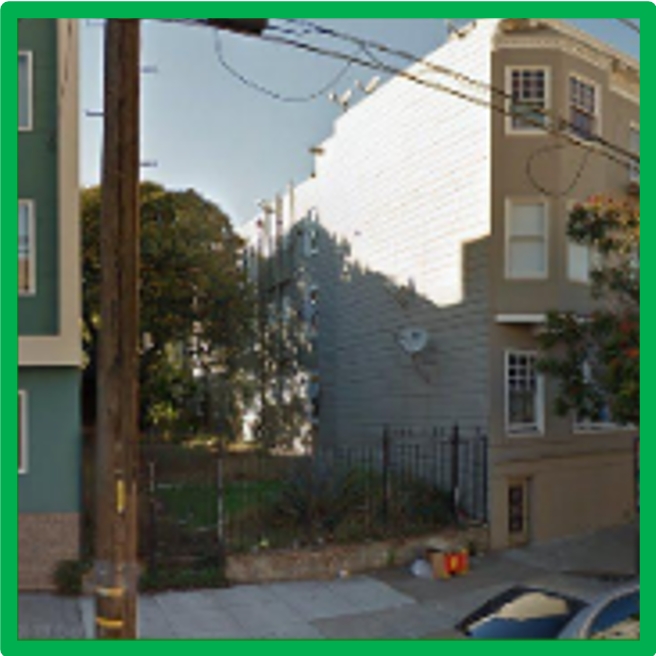}&\includegraphics[width=0.1\textwidth]{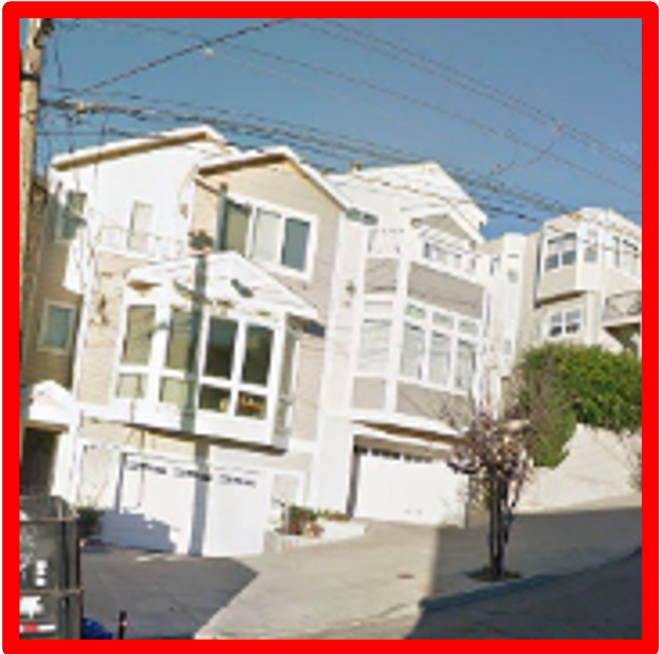}&\includegraphics[width=0.1\textwidth]{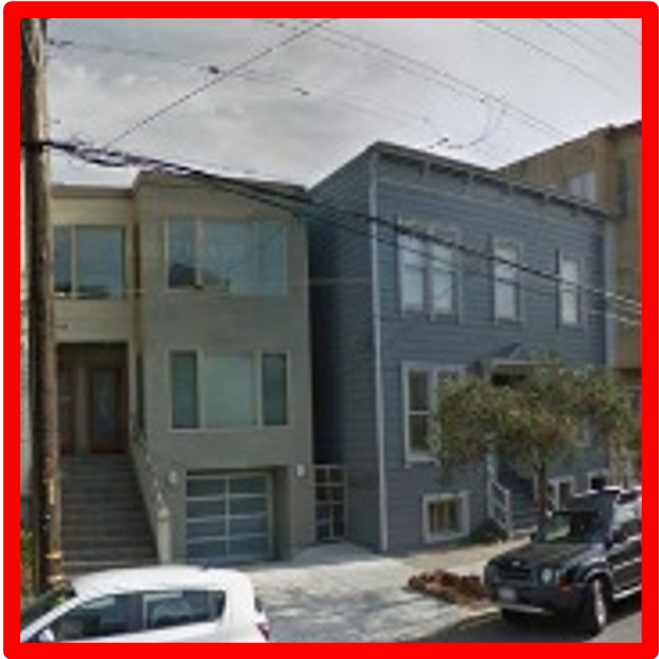}&\includegraphics[width=0.1\textwidth]{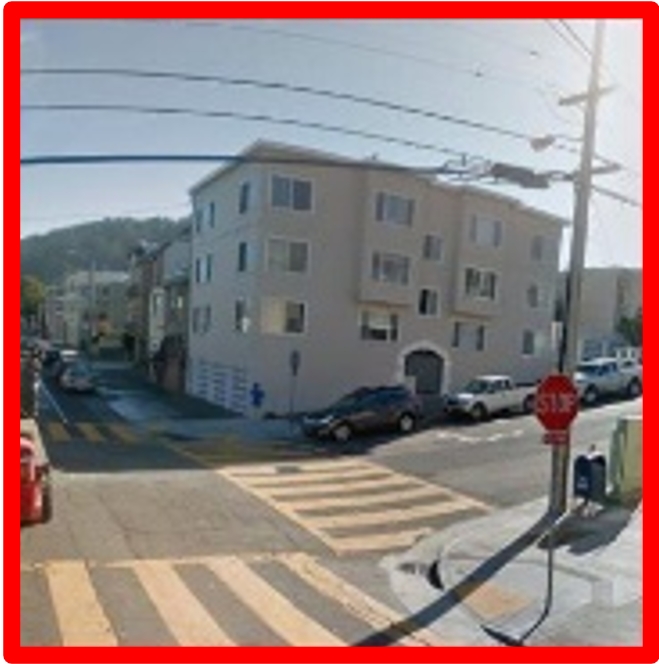}&\includegraphics[width=0.1\textwidth]{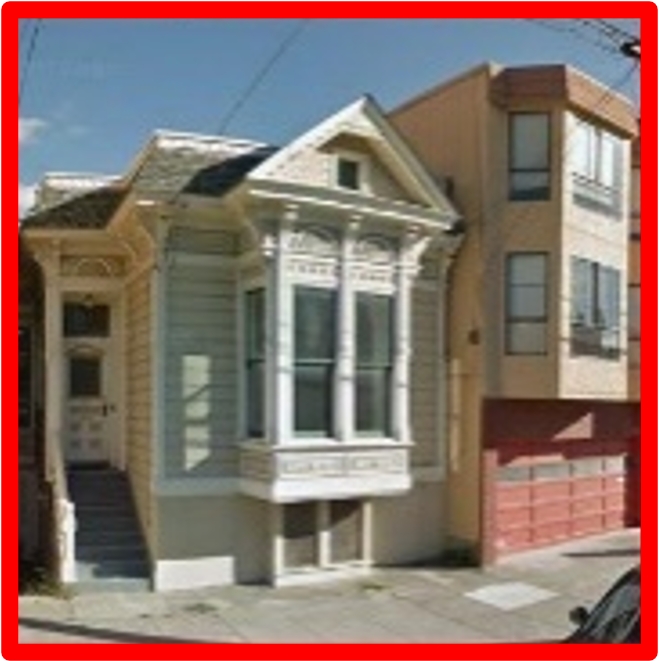} \\
    
    &\includegraphics[width=0.1\textwidth]{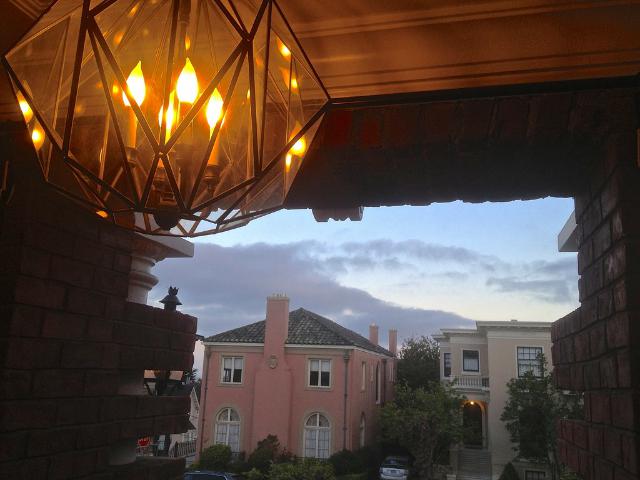}& \includegraphics[width=0.1\textwidth]{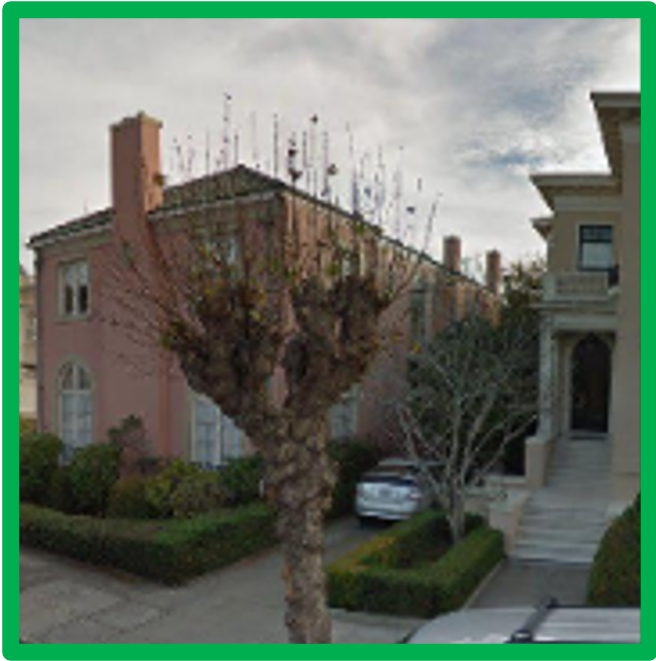}&\includegraphics[width=0.1\textwidth]{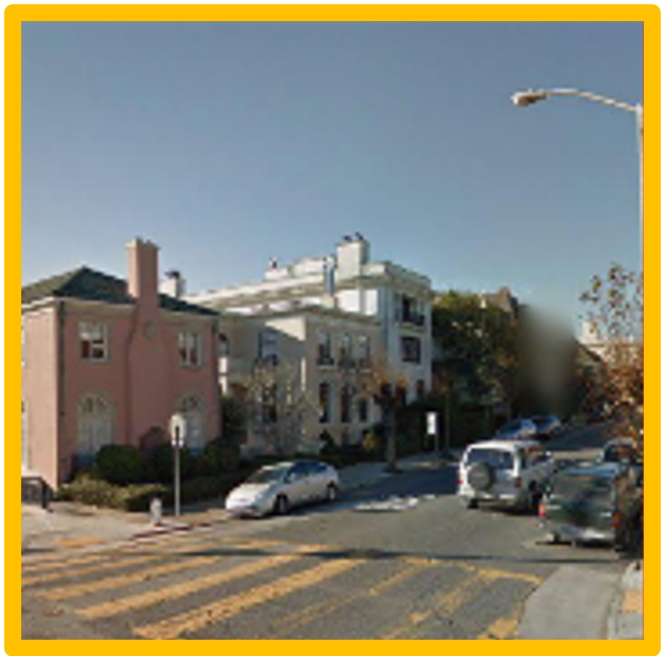}&\includegraphics[width=0.1\textwidth]{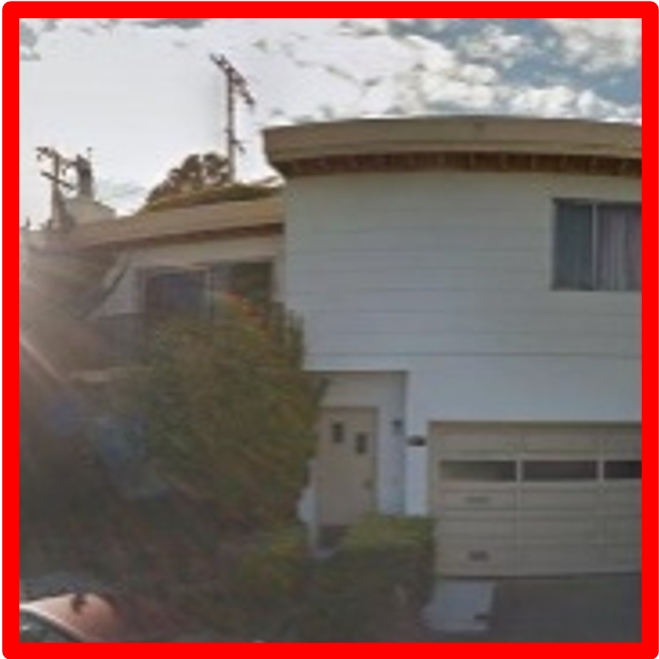}&\includegraphics[width=0.1\textwidth]{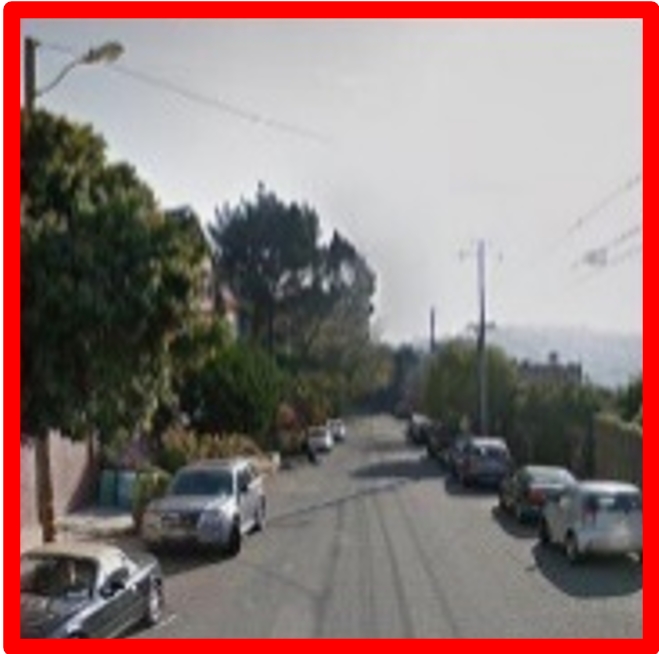}&\includegraphics[width=0.1\textwidth]{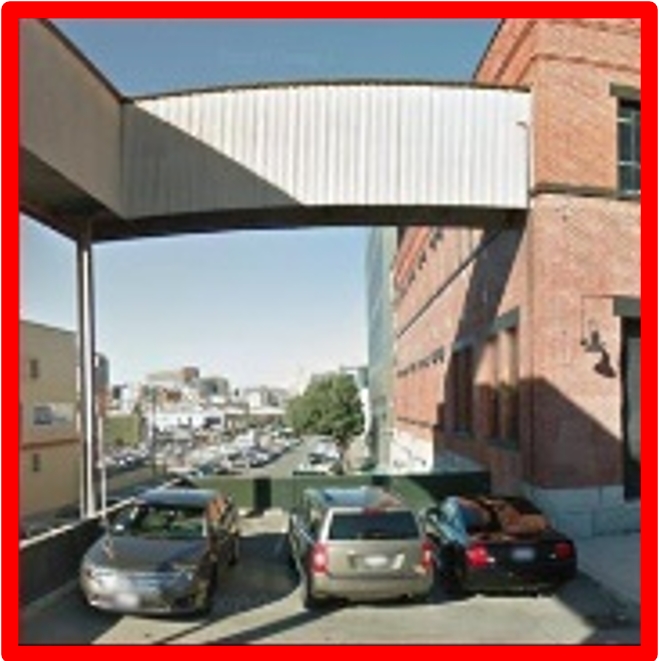} \\
    \hline
    Season Change &\includegraphics[width=0.1\textwidth]{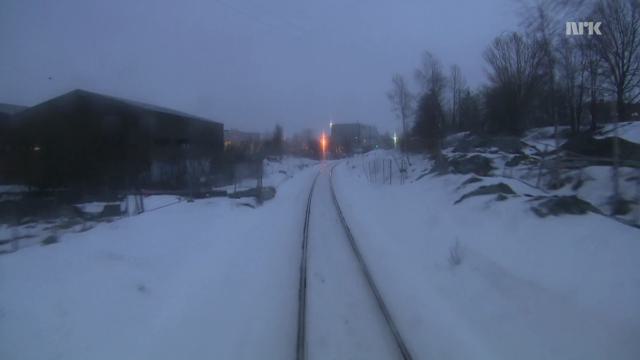}& \includegraphics[width=0.1\textwidth]{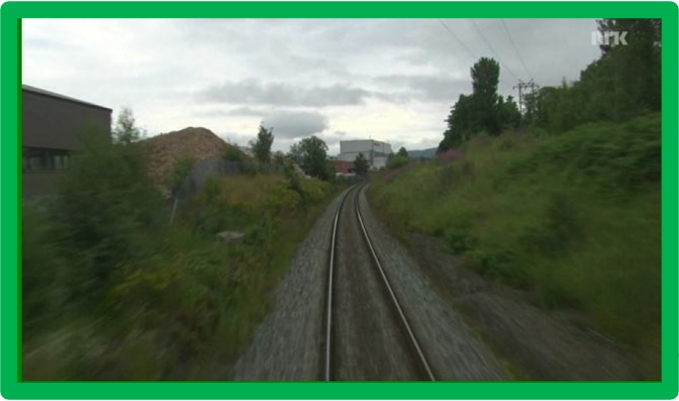}&\includegraphics[width=0.1\textwidth]{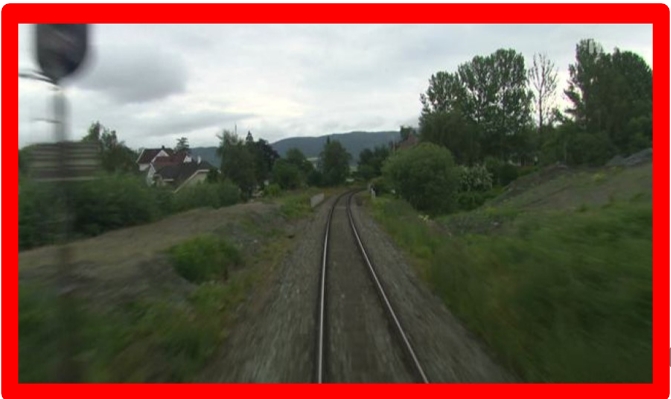}&\includegraphics[width=0.1\textwidth]{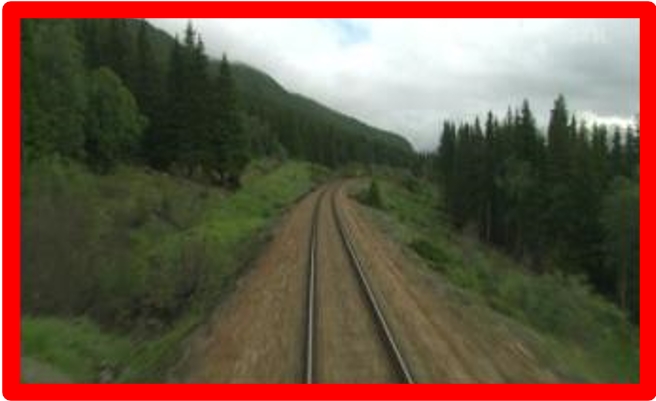}&\includegraphics[width=0.1\textwidth]{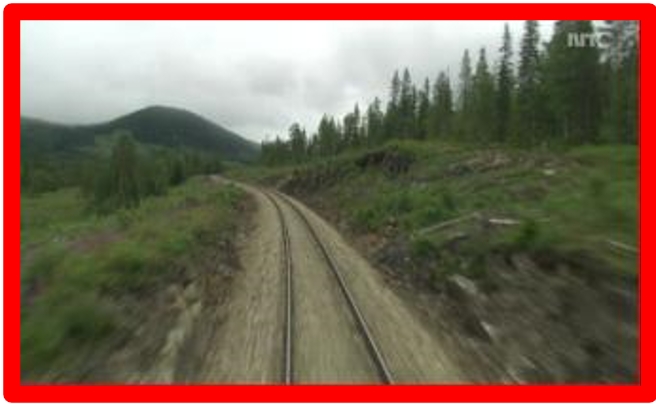}&\includegraphics[width=0.1\textwidth]{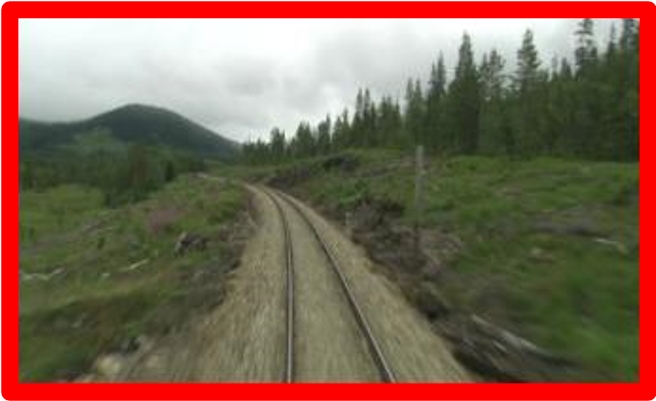} \\
    
    &\includegraphics[width=0.1\textwidth]{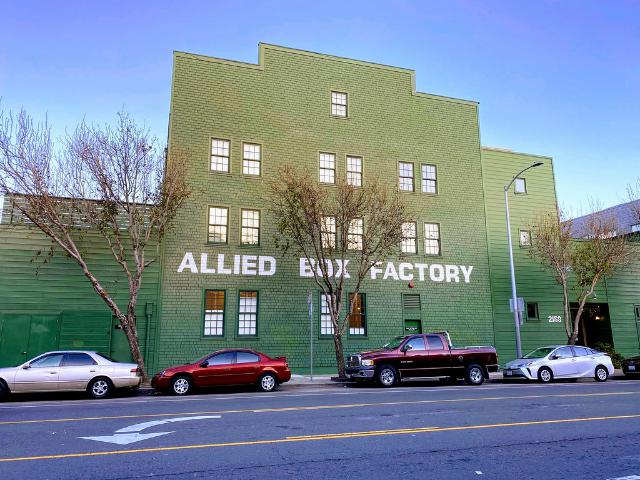}& \includegraphics[width=0.1\textwidth]{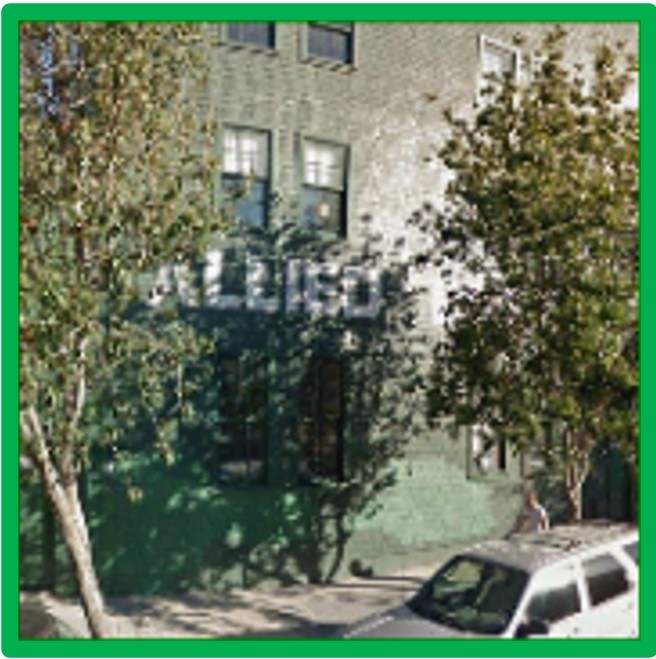}&\includegraphics[width=0.1\textwidth]{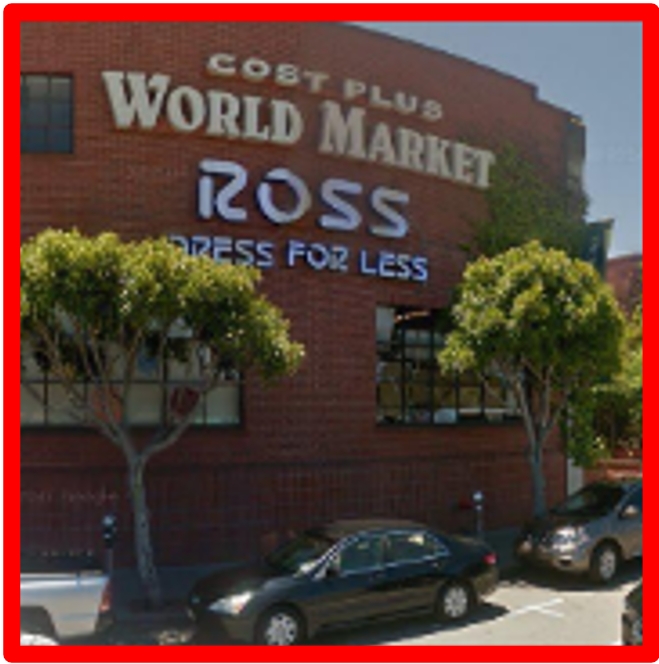}&\includegraphics[width=0.1\textwidth]{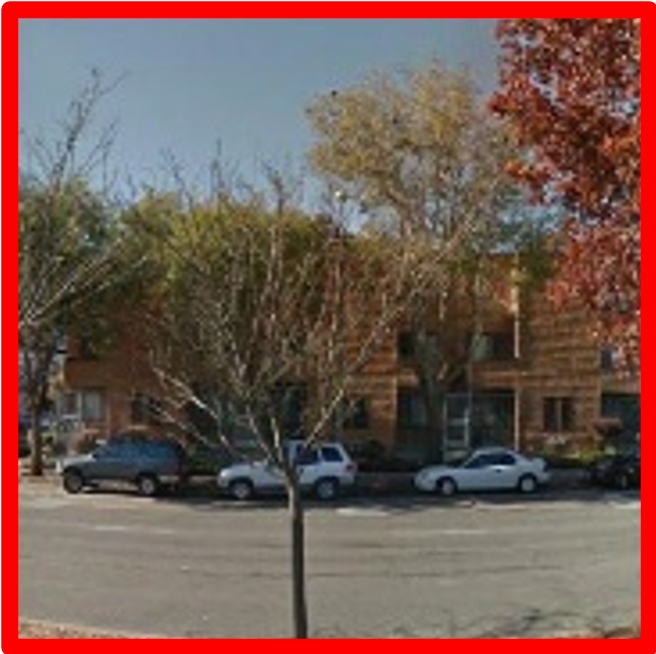}&\includegraphics[width=0.1\textwidth]{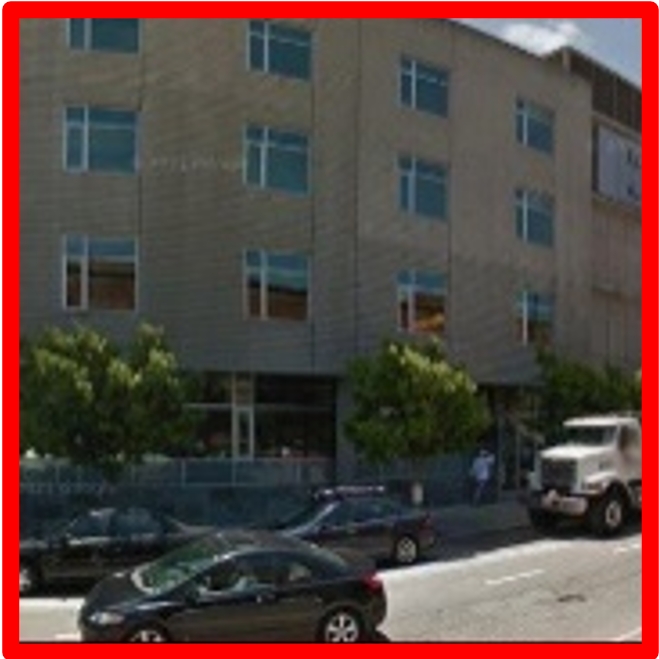}&\includegraphics[width=0.1\textwidth]{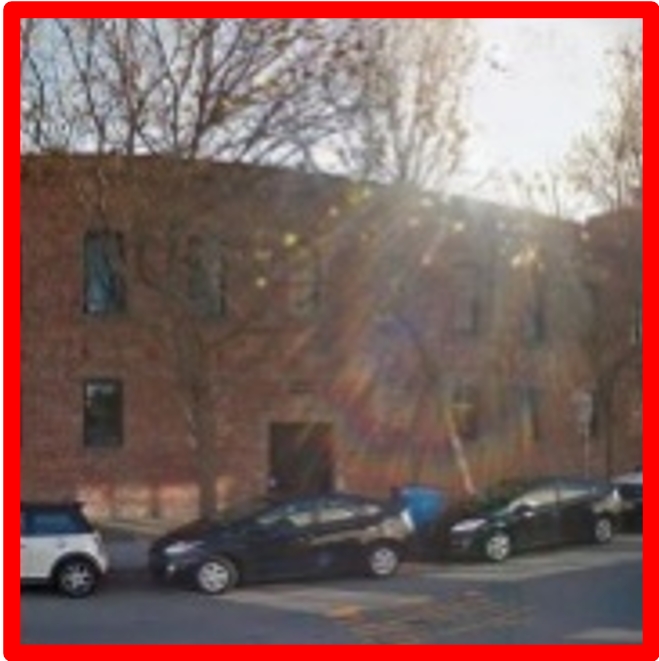} \\
    \hline
    \end{tabular}
    \label{tab:Comparison of image retrieval}
    \end{table*}
    
    In this study, we compared the performance of DINO-Mix with those of existing SOTA methods, including MixVPR, NetVLAD, ConvAP, and CosPlace, in image retrieval tasks. To demonstrate the robustness of DINO-Mix for VPR in complex environments, we selected several representative image retrieval cases from the Tokyo24/7, SF-XL-Testv1, and Nordland datasets. We presented four challenging scenarios: viewpoint changes, illumination changes, object occlusions, and seasonal variations. In cases where DINO-Mix succeeded, the other methods failed to accurately locate the query image, as displayed in Tab.\ref{tab:Comparison of image retrieval}.
    
    \textbf{Viewpoint Change:} Viewpoint changes encompass variations in the field angle and field range, posing challenges for image retrieval. Rows 1 and 2 in Tab.\ref{tab:Comparison of image retrieval} show examples of viewpoint changes in terms of field angle and field range. Notably, only DINO-Mix resists interference caused by viewpoint changes and retrieves the correct image, whereas the other methods retrieve similar buildings or scenes.

    \textbf{Illumination Change:} Illumination changes significantly affect image retrieval accuracy. Dim lighting conditions can blur textures in images, adversely affecting feature extraction and, consequently, image retrieval accuracy. Rows 3 and 4 in Tab.\ref{tab:Comparison of image retrieval} depict the image retrieval cases under dark conditions. Rows 5 and 6 present nighttime scenarios with artificial and natural light variations, respectively. DINO-Mix exhibits strong robustness against illumination changes, whereas the other methods suffer from the effects of lighting variations and fail to retrieve accurate results.

    \begin{table*}[!t]
    \renewcommand{\thetable}{5}
    \caption{\emph{\textbf{The attention map visualization of the query images.}}}
    \centering
    \begin{tabular}{ c p{2cm} c p{2cm} c p{2cm} c p{2cm} c p{2cm} c p{2cm}}
    \hline
    Query&DINO-Mix&MixVPR&NetVLAD&ConvAP&CosPlace\\
    \hline
 \includegraphics[width=0.1\textwidth]{Pics/q1.jpg}& \includegraphics[width=0.1\textwidth]{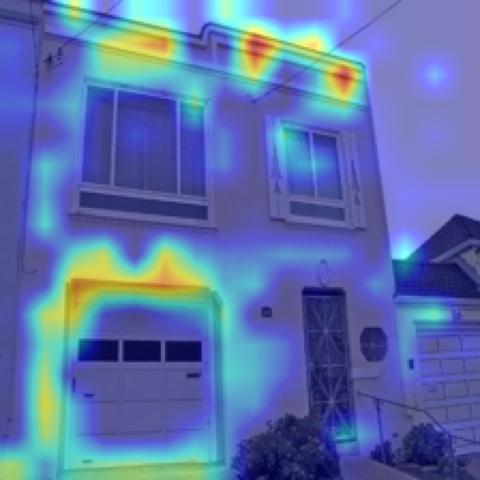}&\includegraphics[width=0.1\textwidth]{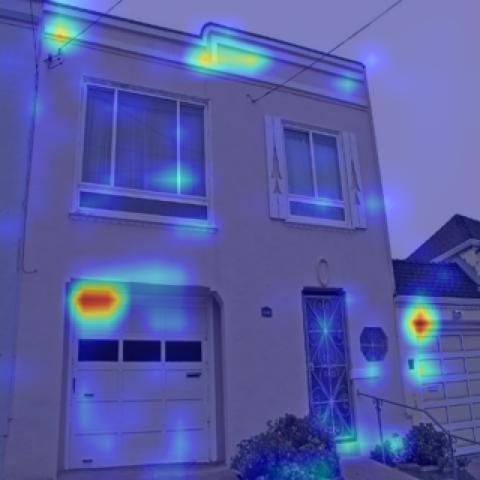}&\includegraphics[width=0.1\textwidth]{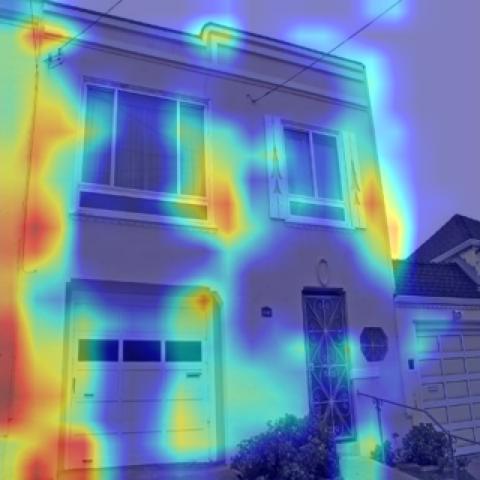}&\includegraphics[width=0.1\textwidth]{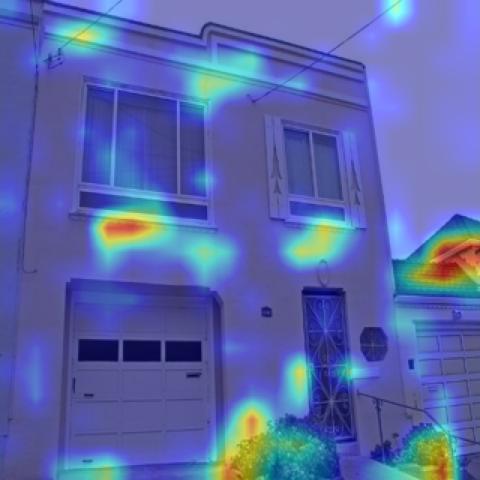}&\includegraphics[width=0.1\textwidth]{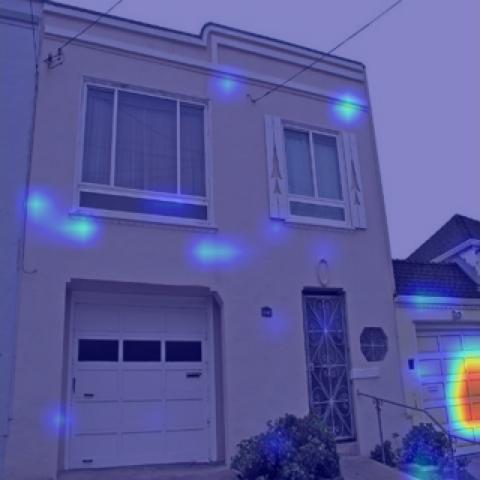} \\
 \includegraphics[width=0.06\textwidth]{Pics/q2.jpg}& \includegraphics[width=0.1\textwidth]{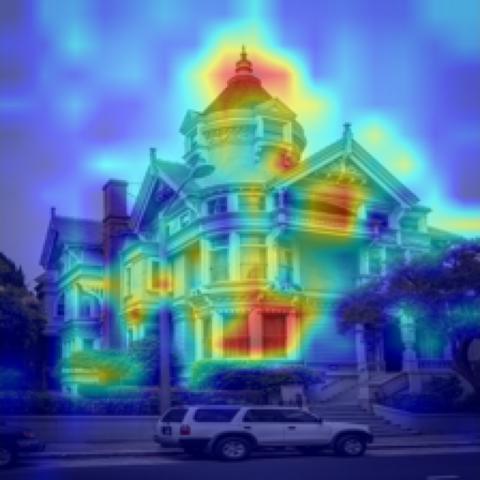}&\includegraphics[width=0.1\textwidth]{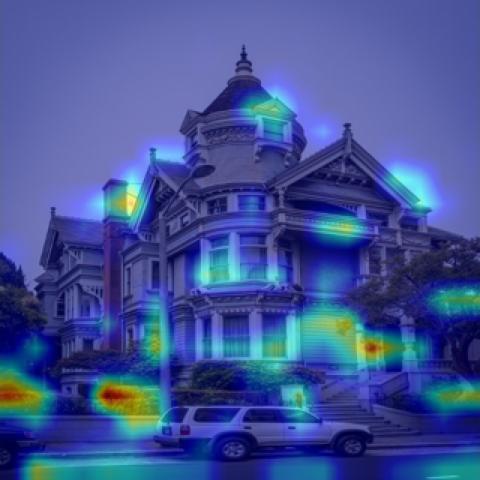}&\includegraphics[width=0.1\textwidth]{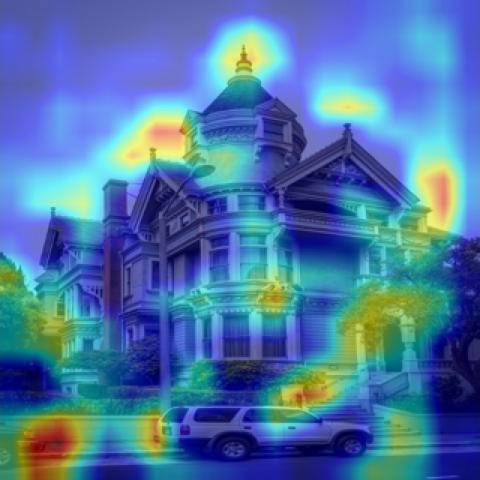}&\includegraphics[width=0.1\textwidth]{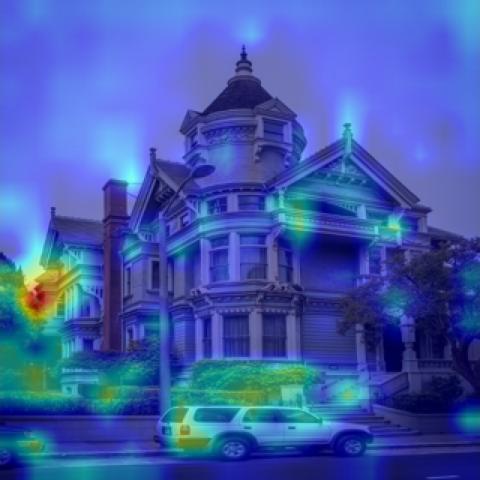}&\includegraphics[width=0.1\textwidth]{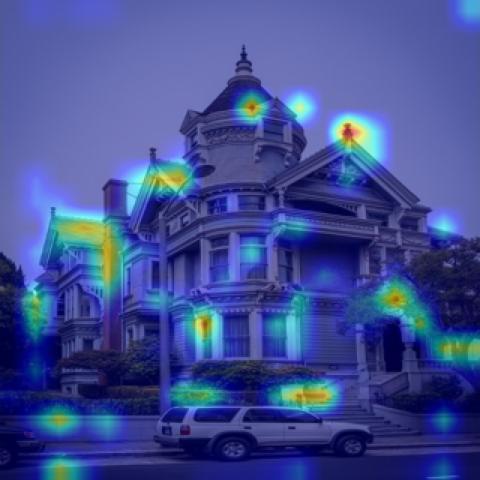} \\
 \includegraphics[width=0.073\textwidth]{Pics/q3.jpg}& \includegraphics[width=0.1\textwidth]{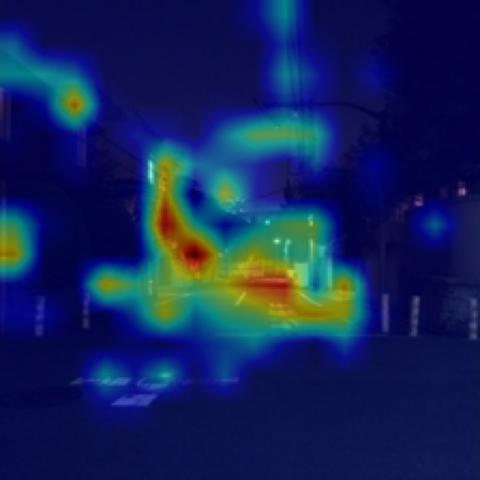}&\includegraphics[width=0.1\textwidth]{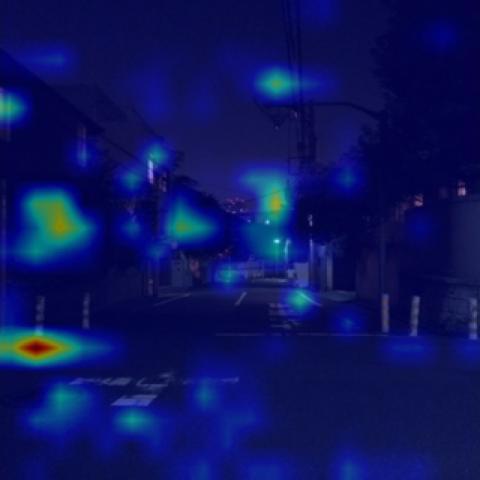}&\includegraphics[width=0.1\textwidth]{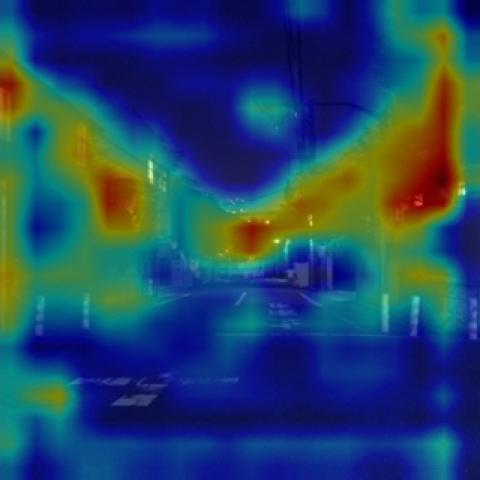}&\includegraphics[width=0.1\textwidth]{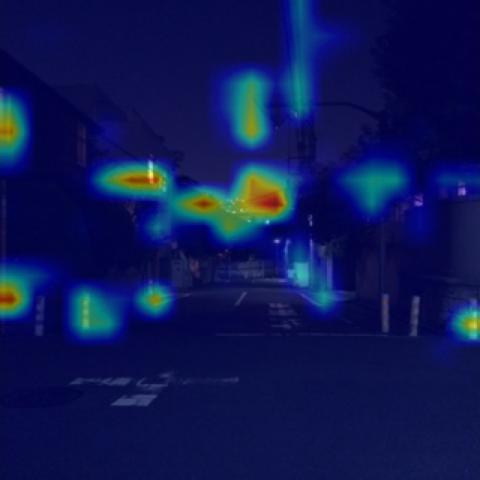}&\includegraphics[width=0.1\textwidth]{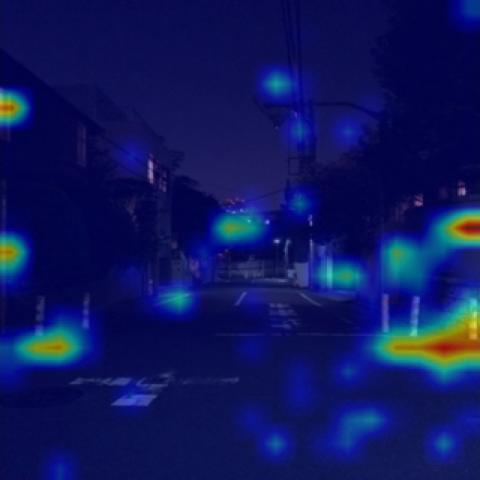} \\
 \includegraphics[width=0.073\textwidth]{Pics/q4.jpg}& \includegraphics[width=0.1\textwidth]{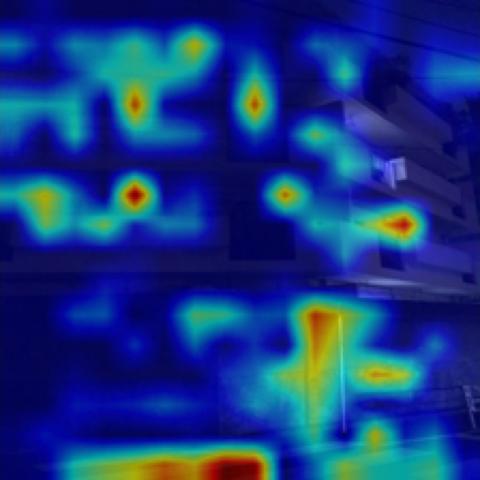}&\includegraphics[width=0.1\textwidth]{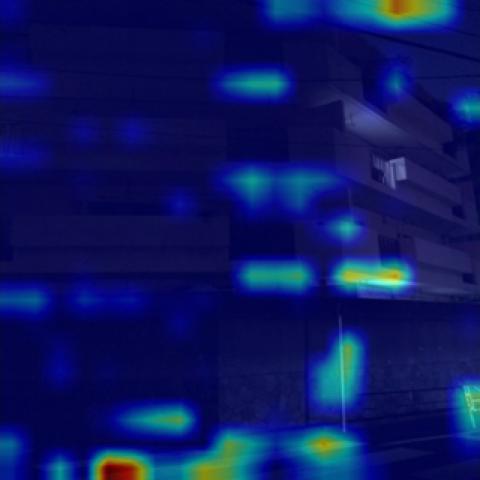}&\includegraphics[width=0.1\textwidth]{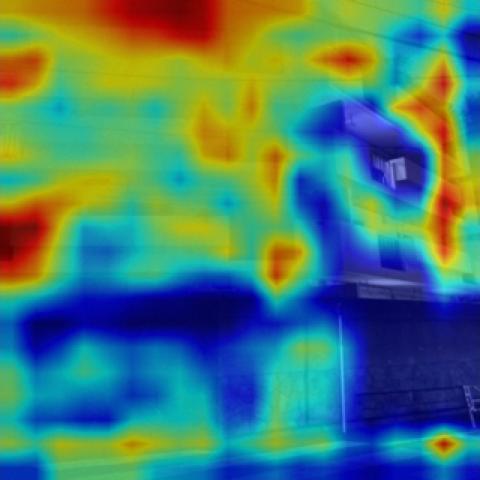}&\includegraphics[width=0.1\textwidth]{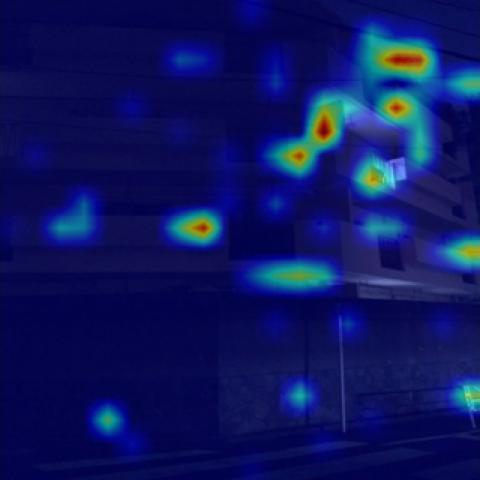}&\includegraphics[width=0.1\textwidth]{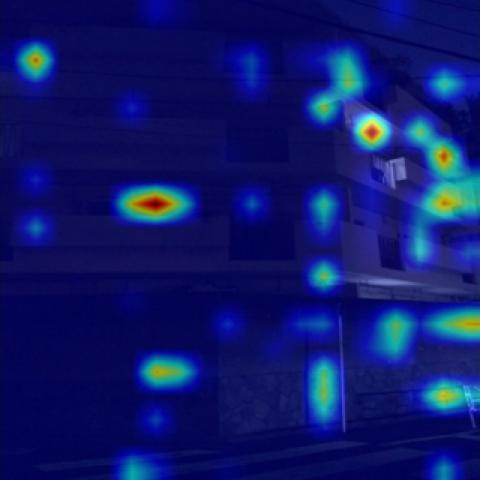} \\
 \includegraphics[width=0.1\textwidth]{Pics/q5.jpg}& \includegraphics[width=0.1\textwidth]{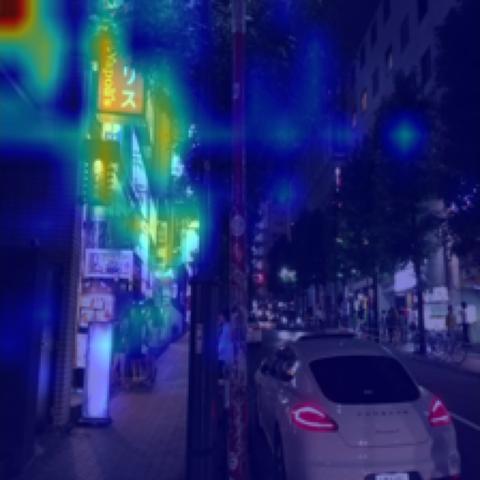}&\includegraphics[width=0.1\textwidth]{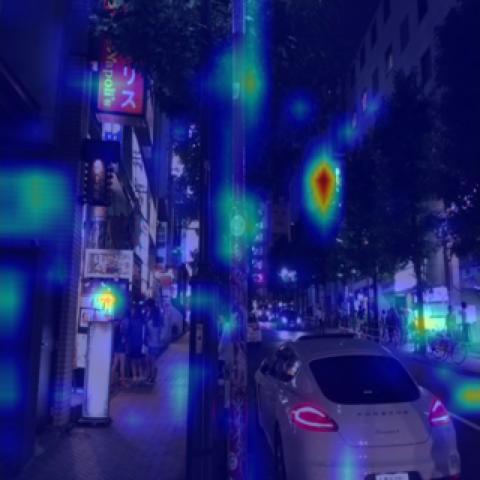}&\includegraphics[width=0.1\textwidth]{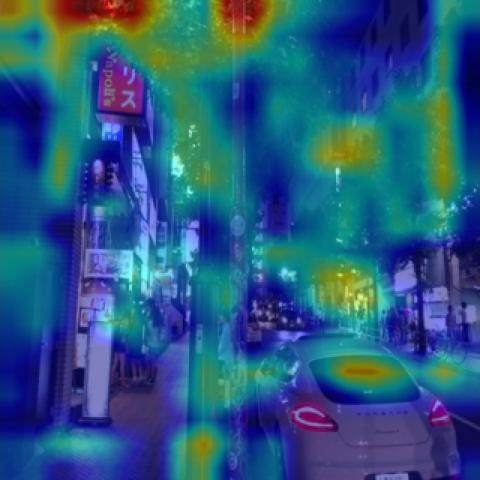}&\includegraphics[width=0.1\textwidth]{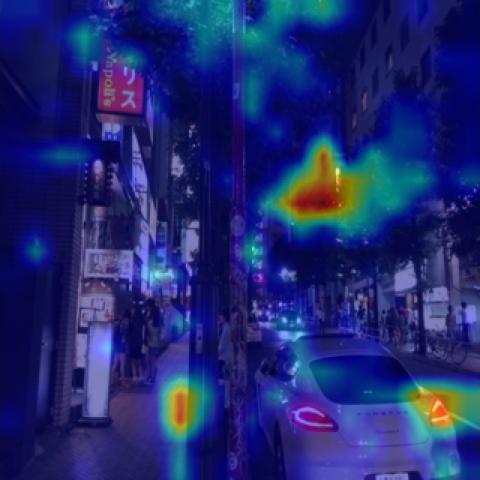}&\includegraphics[width=0.1\textwidth]{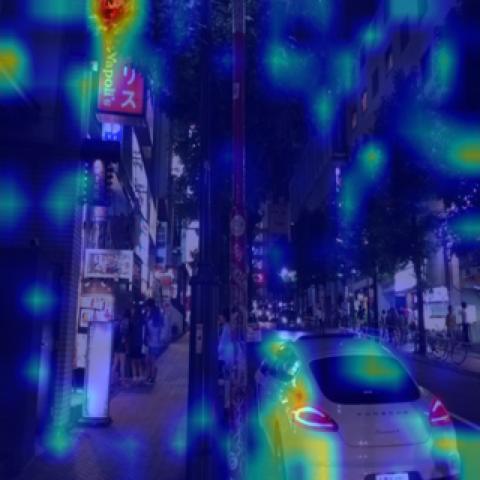} \\
 \includegraphics[width=0.07\textwidth]{Pics/q6.jpg}& \includegraphics[width=0.1\textwidth]{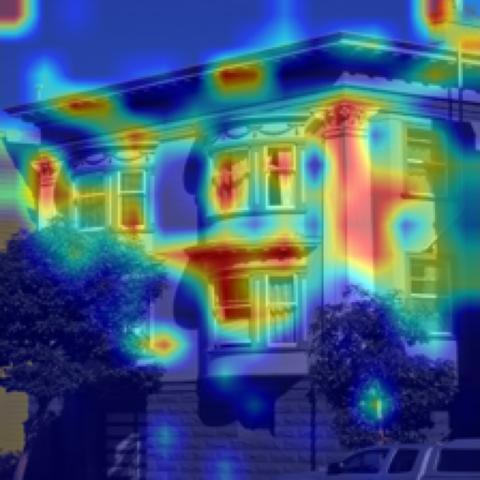}&\includegraphics[width=0.1\textwidth]{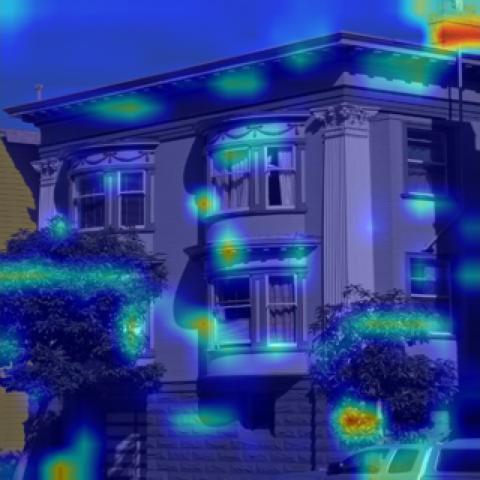}&\includegraphics[width=0.1\textwidth]{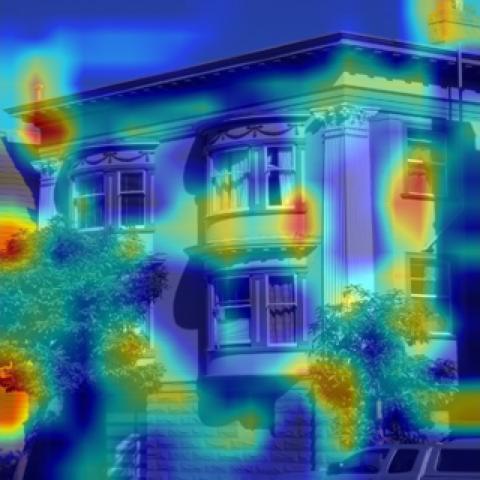}&\includegraphics[width=0.1\textwidth]{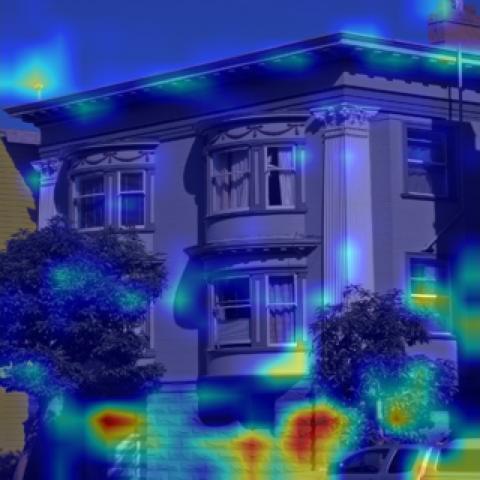}&\includegraphics[width=0.1\textwidth]{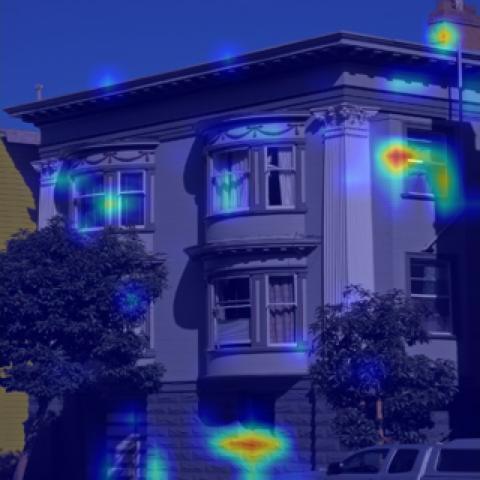} \\
 \includegraphics[width=0.07\textwidth]{Pics/q7.jpg}& \includegraphics[width=0.1\textwidth]{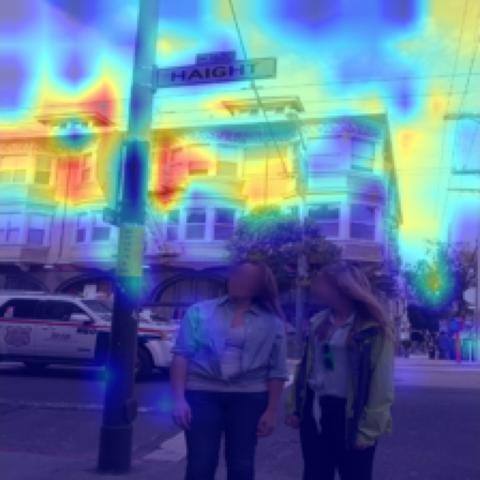}&\includegraphics[width=0.1\textwidth]{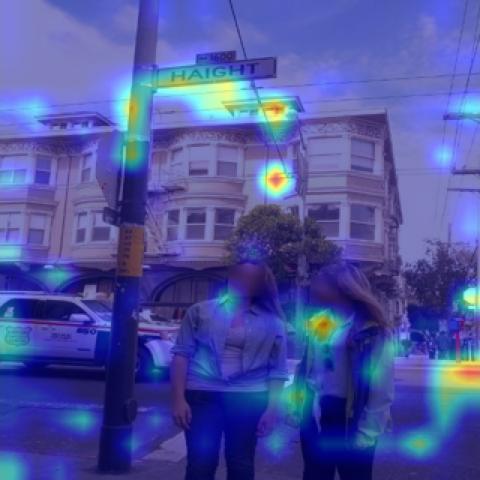}&\includegraphics[width=0.1\textwidth]{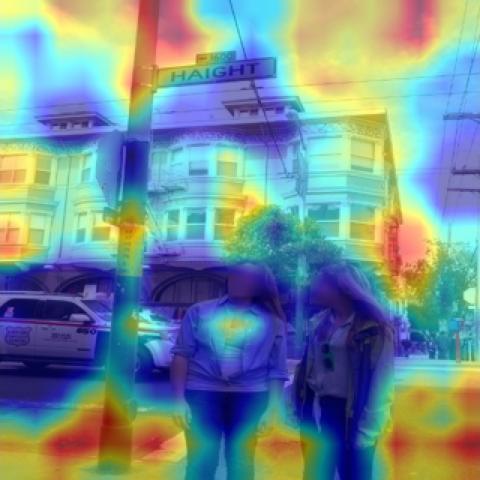}&\includegraphics[width=0.1\textwidth]{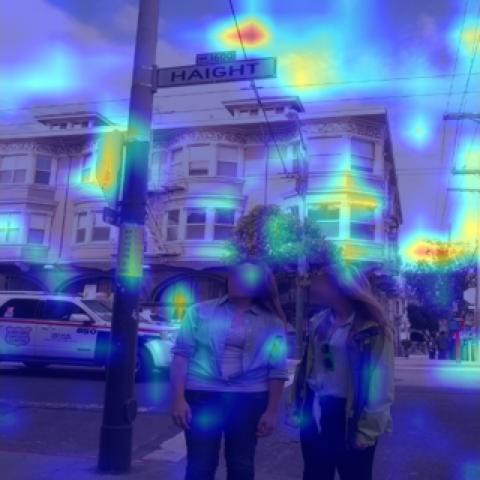}&\includegraphics[width=0.1\textwidth]{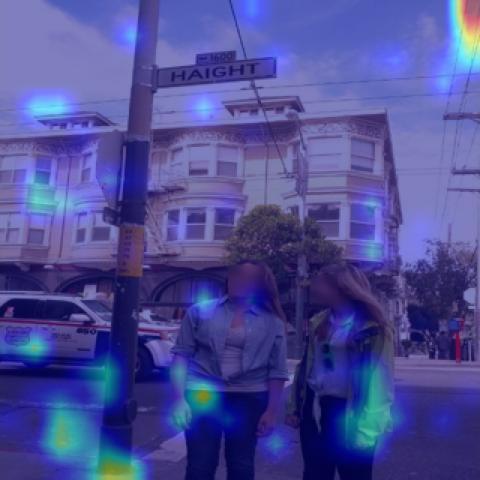} \\
 \includegraphics[width=0.07\textwidth]{Pics/q8.jpg}& \includegraphics[width=0.1\textwidth]{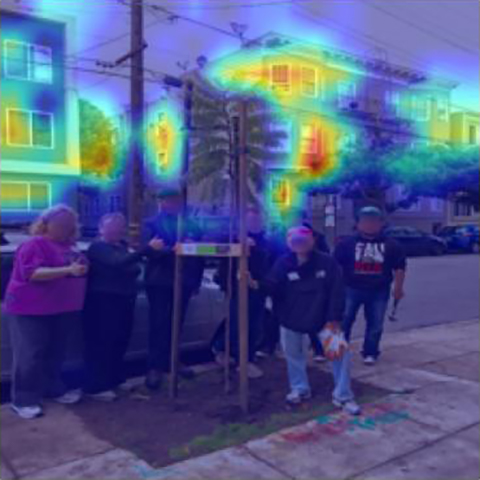}&\includegraphics[width=0.1\textwidth]{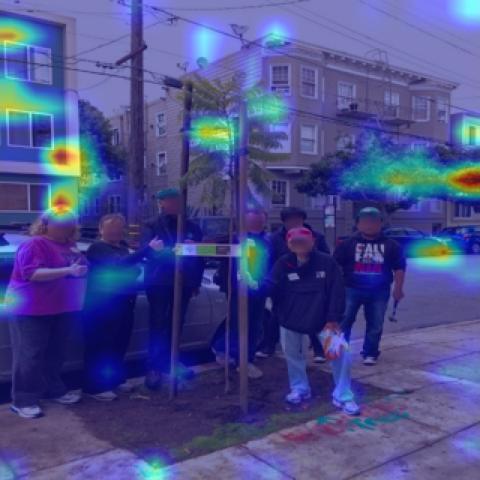}&\includegraphics[width=0.1\textwidth]{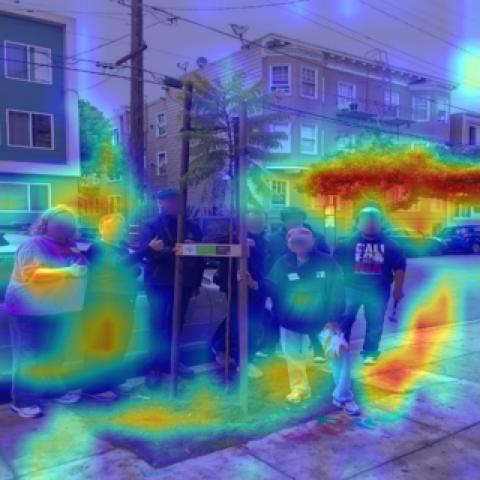}&\includegraphics[width=0.1\textwidth]{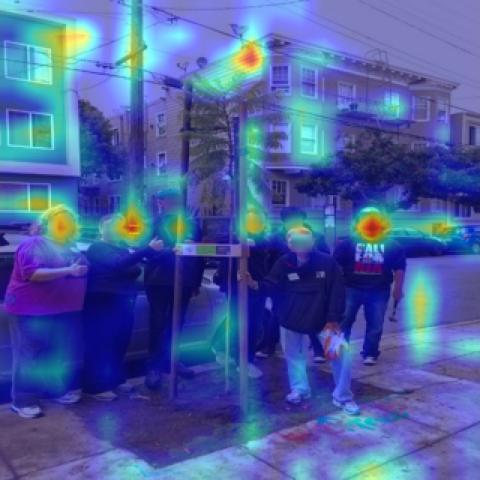}&\includegraphics[width=0.1\textwidth]{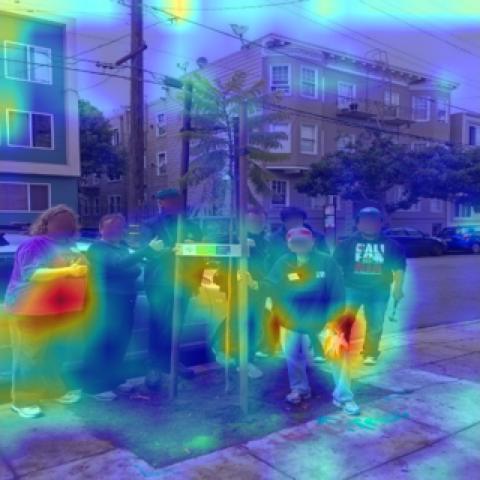} \\
 \includegraphics[width=0.1\textwidth]{Pics/q9.jpg}& \includegraphics[width=0.1\textwidth]{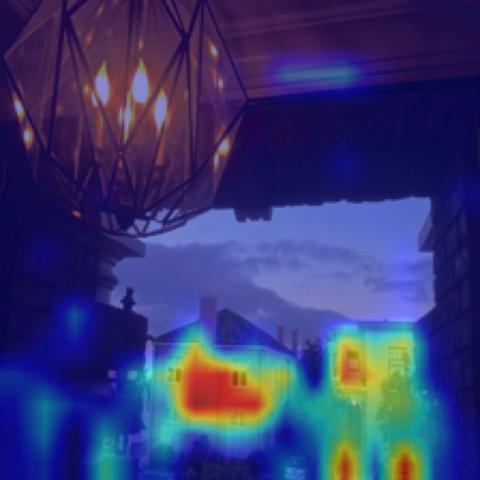}&\includegraphics[width=0.1\textwidth]{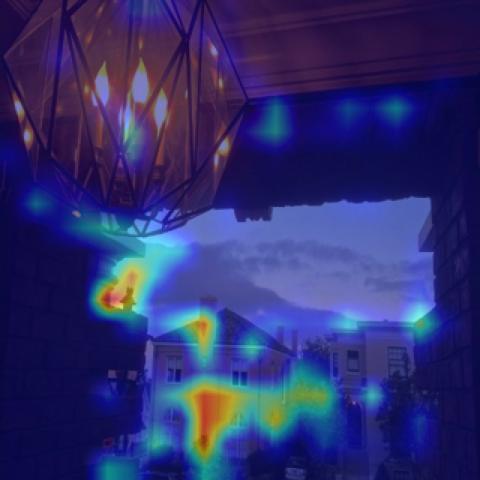}&\includegraphics[width=0.1\textwidth]{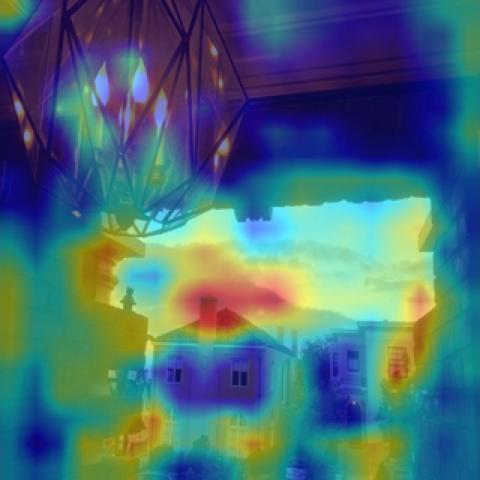}&\includegraphics[width=0.1\textwidth]{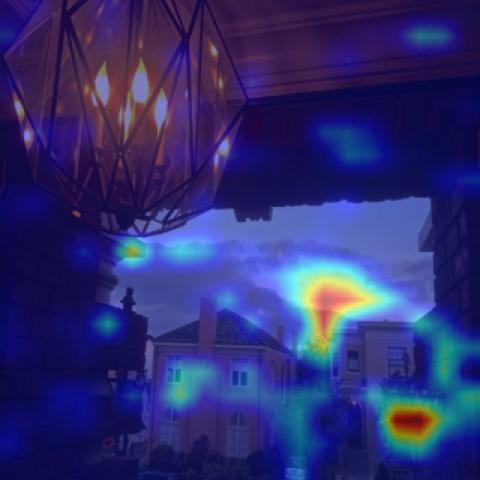}&\includegraphics[width=0.1\textwidth]{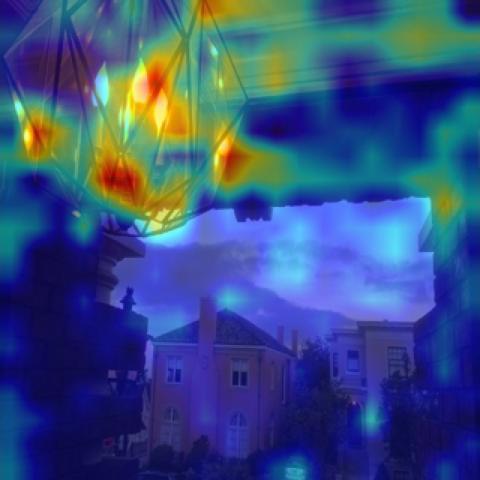} \\
 \includegraphics[width=0.1\textwidth]{Pics/q10.jpg}& \includegraphics[width=0.1\textwidth]{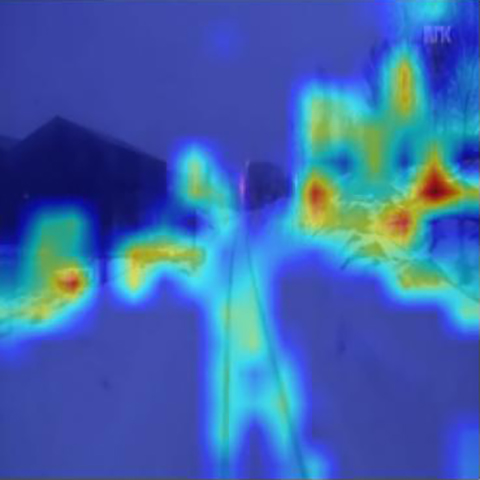}&\includegraphics[width=0.1\textwidth]{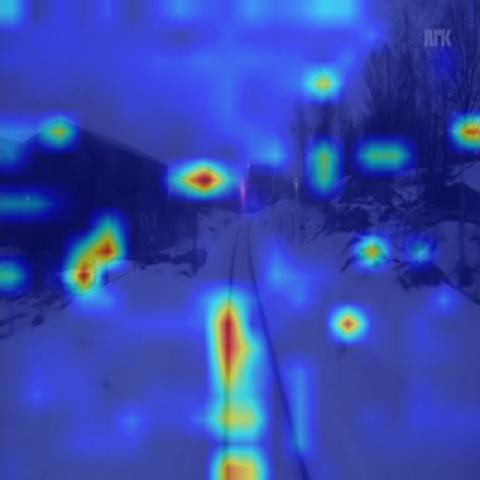}&\includegraphics[width=0.1\textwidth]{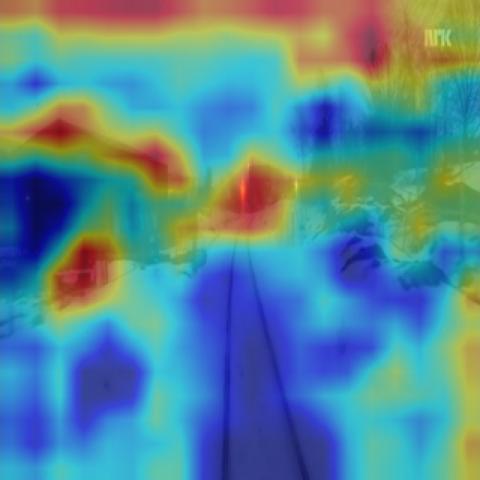}&\includegraphics[width=0.1\textwidth]{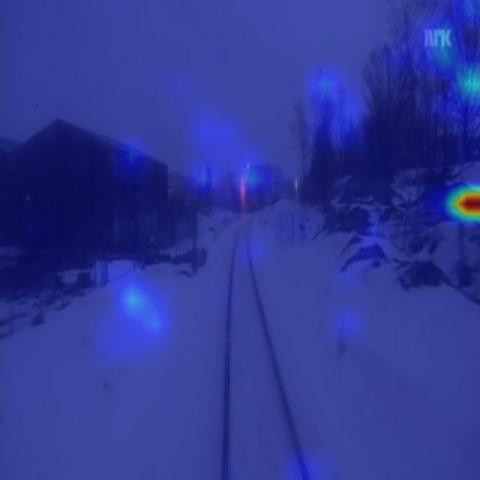}&\includegraphics[width=0.1\textwidth]{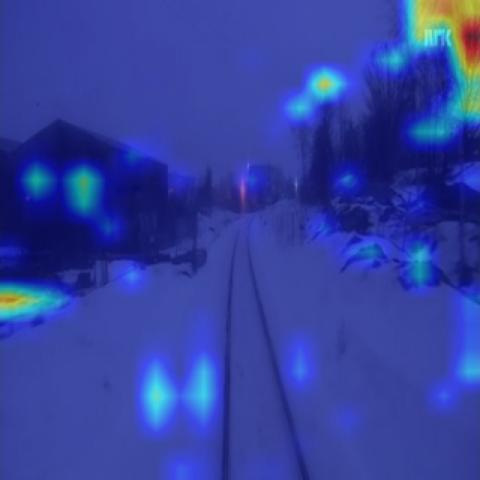} \\
 \includegraphics[width=0.1\textwidth]{Pics/q11.jpg}& \includegraphics[width=0.1\textwidth]{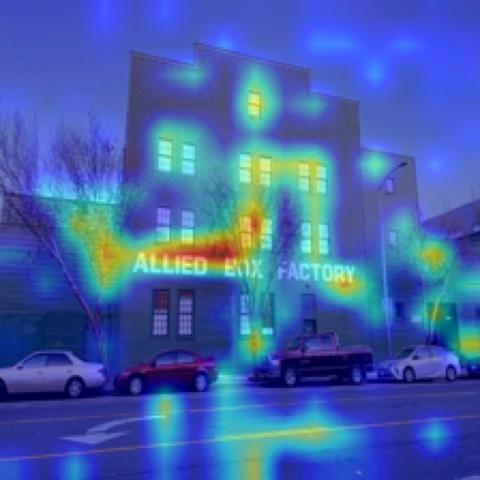}&\includegraphics[width=0.1\textwidth]{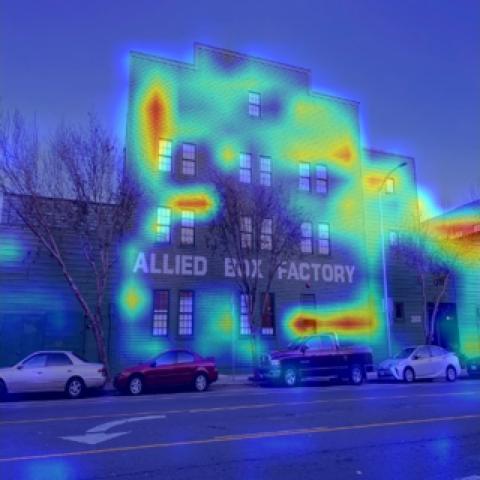}&\includegraphics[width=0.1\textwidth]{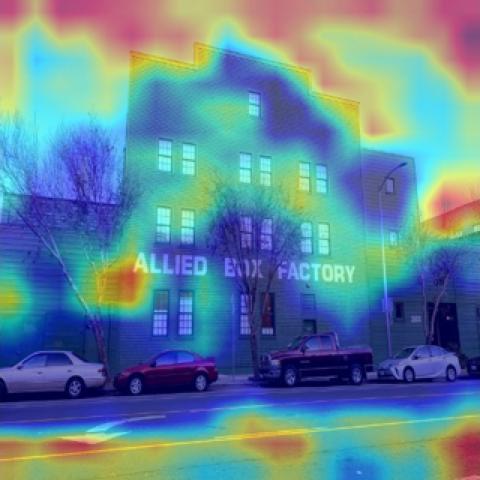}&\includegraphics[width=0.1\textwidth]{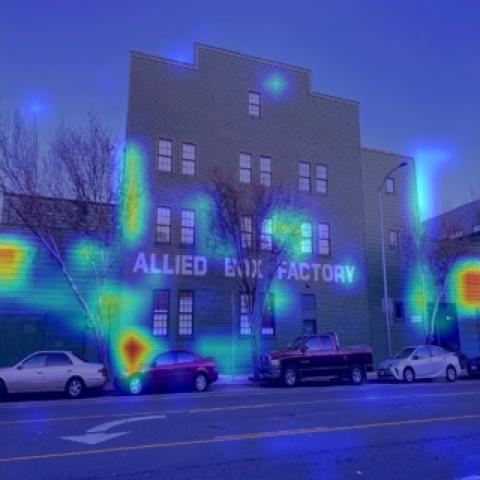}&\includegraphics[width=0.1\textwidth]{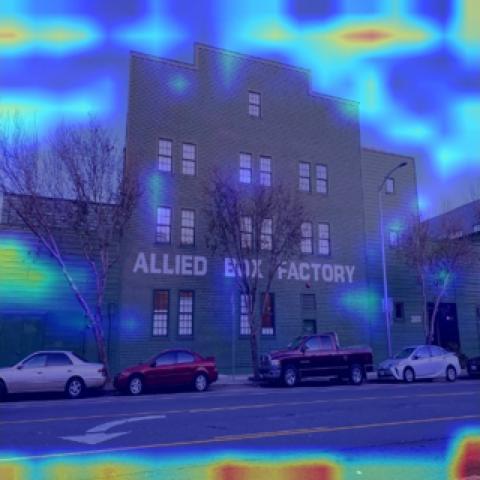} \\
    \hline
    \end{tabular}
    \label{tab:attention map visualization}
    \end{table*}
    
    \textbf{Occlusion:} Image retrieval focuses primarily on objects, such as buildings, facilities, and natural landscapes. However, pedestrians, vehicles, and other objects can interfere with the semantic information in an image, posing challenges for image retrieval. As shown in rows 7 and 8 in Tab.\ref{tab:Comparison of image retrieval}, where a large number of pedestrians are present in the query images, and in row 9, where the influence of buildings is significant, these occlusions pose significant difficulties for image retrieval. MixVPR retrieved the correct content but exceeded the threshold $s$ ($25$ $m$) in the localization results. In contrast, DINO-Mix successfully extracted the correct features from the images and retrieved accurate results despite these challenges.
    
    \textbf{Season Change:} The appearance characteristics of a location undergo significant changes in different seasons, such as heavy snowfall in winter (as illustrated in row 10 of Tab.\ref{tab:Comparison of image retrieval}), and leaves falling from trees (row 11). These seasonal variations also have a profound impact on the image retrieval accuracy. Under such challenging circumstances, DINO-Mix overcame the drastic contrast caused by seasonal changes and achieved satisfactory results.

 \subsection{Attention map visualization} 
 \label{attention map visualization}

    To provide a more intuitive demonstration of the superiority of DINO-Mix over other VPR methods, we visualized their attention maps as presented in \ref{tab:attention map visualization}. The attention scores are represented by varying colors from blue to green to red, indicating low to high attention levels. Our analysis reveals that DINO-Mix can focus more on buildings, object contours, and textures, which are crucial factors for image retrieval. In contrast, it effectively excludes negative elements such as pedestrians, cars, and occlusions. This suggests that DINO-Mix has a greater ability to capture essential features and extract more robust image representations.

%% file: Tex_content/conclusion.tex
In this paper, we proposed a novel VPR framework, called DINO-Mix. First, we modified and fine-tuned the structure of the DINOv2 model. We then converted the extracted features from the backbone network into feature maps and employed the mixed feature aggregation module to aggregate these feature maps to obtain global feature vectors. The experimental results on various test sets demonstrate that the proposed DINO-Mix model outperforms SOTA methods in terms of VPR accuracy, with an average improvement of $5.14\%$ across test sets containing challenging conditions. Furthermore, through a series of image retrieval examples under difficult circumstances, we demonstrated that the performance of the DINO-Mix architecture significantly surpasses that of current SOTA architectures. We identified that changes in illumination pose a significant challenge. Therefore, our future work will focus on enhancing images with poor lighting conditions through image augmentation techniques to further improve the VPR accuracy.